\pdfoutput=1
\documentclass{article}
\usepackage{arxiv}
\usepackage[utf8]{inputenc}
\usepackage[T1]{fontenc}
\usepackage{times}
\usepackage[round,authoryear]{natbib}
\setcitestyle{authoryear,round,citesep={;},aysep={,},yysep={;}}

\usepackage{amsmath,amsfonts,bm}

\def\eqref#1{equation~\ref{#1}}

\def\1{\bm{1}}

\DeclareMathAlphabet{\mathsfit}{\encodingdefault}{\sfdefault}{m}{sl}
\SetMathAlphabet{\mathsfit}{bold}{\encodingdefault}{\sfdefault}{bx}{n}

\usepackage{amssymb}
\usepackage{amsthm}
\usepackage{booktabs}
\usepackage{multirow}
\usepackage{graphicx}
\usepackage{capt-of}
\usepackage{xcolor}
\usepackage{pifont}
\usepackage{makecell}
\usepackage{algorithm}
\usepackage{algpseudocode}
\usepackage[hidelinks]{hyperref}
\usepackage{url}

\newtheorem{proposition}{Proposition}

\newtheorem{assumption}{Assumption}
\theoremstyle{remark}\newtheorem{remark}{Remark}

\newcommand{\gc}[1]{\textcolor{gray}{#1}}

\newlength{\figw}

\renewcommand{\shorttitle}{AutoScaffold: AI-Guided Scaffolding for RL of Language Models}
\renewcommand{\headeright}{Preprint}
\renewcommand{\undertitle}{Preprint}

\title{AutoScaffold: AI-Guided Scaffolding for \\ Reinforcement Learning of Language Models}

\author{%
  Zijian Zhang$^{1}$ \quad Rizhen Hu$^{1}$ \quad Jiangweizhi Peng$^{1}$ \quad Shiyang Li$^{1}$ \\
  \textbf{Sujay Bhatt$^{2}$ \quad Sihan Zeng$^{2}$ \quad Mingyi Hong$^{1}$} \\[4pt]
  {\normalfont $^{1}$University of Minnesota, Twin Cities \qquad $^{2}$JPMorgan AI Research}
}
\date{}

\begin{document}
\maketitle

\vspace{-0.46in}
\begin{center}
\setlength{\abovecaptionskip}{5pt}
\includegraphics[width=\figw]{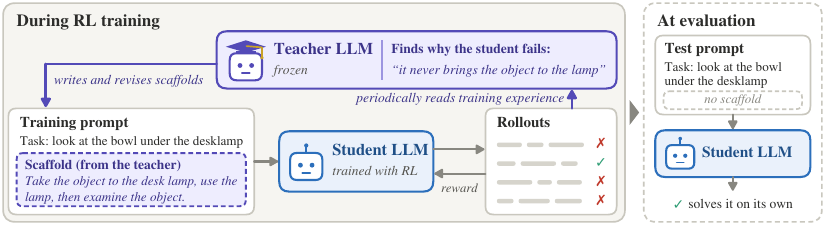}
\captionof{figure}{\textbf{A teacher LLM helps a student LLM learn in AutoScaffold.} During RL training, the frozen teacher reads the student's training experience and writes scaffolds into its training prompts. Scaffolds are removed at evaluation.}
\label{fig:hero}
\end{center}
\vspace{0.12in}

\begin{abstract}
Reinforcement learning with verifiable rewards (RLVR) produces rich training experience: the model's reasoning, its actions, and the feedback it receives. Both successful and failed attempts can reveal useful strategies, recurring mistakes, and inefficient behavior, yet outcome rewards reduce each attempt to a single number and leave most of this information unused. Training-time scaffolds, such as strategies or hints added to the prompt during training, offer a natural way to turn this information into guidance for the model. In practice, however, their design and adaptation are largely left to humans and tailored to specific domains, rather than derived from the model's own training experience.

We introduce \textbf{AutoScaffold}, a unified AI-for-AI framework that turns a student's training experience into adaptive scaffolds, with minimal domain-specific information and no human intervention during training. The goal is to help the student learn more effectively with scaffolds during RL training, so that it performs better on its own when the scaffolds are removed at evaluation.
A frozen teacher periodically examines this experience, diagnoses bottlenecks, and designs scaffolds in two forms: \emph{(i)} strategies shared across a class of tasks and \emph{(ii)} hints for individual hard tasks. The teacher adjusts the content and amount of support as the student learns, and the resulting training experience informs the next round of diagnosis and revision.
We apply the same framework to ALFWorld, search-based QA, mathematical reasoning, and GPU kernel generation. When evaluated without scaffolds, students trained with AutoScaffold outperform those trained with the corresponding RL baselines in all four domains. Gains include 7.2 points in ALFWorld success and 4.0 points in average QA exact match. On GPU kernel generation, AutoScaffold improves both correctness and execution efficiency, increasing the fraction of tasks with a correct kernel that runs faster than the PyTorch reference by up to 5.0 points. On math, our student surpasses QuestA's reported accuracy using approximately one-third as many student training steps, starting from the same partial solutions as QuestA but without its fixed curriculum. Analyses show that the teacher identifies a different failure in each task class from the student's trajectories and shifts its scaffolds from step-by-step workflows to targeted fixes as the student improves, and that its revised scaffolds help the student learn more than keeping the previous scaffold or using no scaffold. These results suggest that a teacher can turn training experience into temporary guidance that improves the student's independent capabilities.
\end{abstract}
\newpage

\section{Introduction}
\label{sec:intro}
Reinforcement learning with verifiable rewards (RLVR) has become a standard approach to post-training language models for reasoning and agentic tasks~\citep{shao2024deepseekmathpushinglimitsmathematical,yu2025dapoopensourcellmreinforcement,feng2025gigpo}. Every training step produces rich experience: the model's reasoning, its actions, and the feedback it receives. Yet RL learns from this experience only through outcome rewards, which reduce each rollout to a single number. Much of what the rollouts reveal is lost: a failed attempt may contain useful reasoning or actions, a successful one may rely on an inefficient strategy, and on tasks that are too hard for the model, nearly every attempt fails and the reward provides no signal at all (Appendix~\ref{sec:theory}). Making better use of training experience is therefore an opportunity to improve post-training beyond what rewards alone can teach.

One promising way to turn such experience into guidance is \emph{training-time scaffolding}, which refers to the procedure of adding auxiliary text (a \emph{scaffold}), such as a strategy, hint, or partial solution, to the prompt of the model being trained (the \emph{student})~\citep{zhang2026scafgrposcaffoldedgrouprelative,li2026questaexpandingreasoningcapacity,lu2026skill0incontextagenticreinforcement}. A scaffold conveys what the reward alone cannot: on tasks the student cannot yet solve, it helps the student succeed so that these tasks produce learning signal, and more generally it shows the student better ways to reason and act. Scaffolds serve only to help the student learn during training and are removed at evaluation, where the student must solve tasks on its own.

\begin{figure}[t]
\centering
\includegraphics[width=\figw]{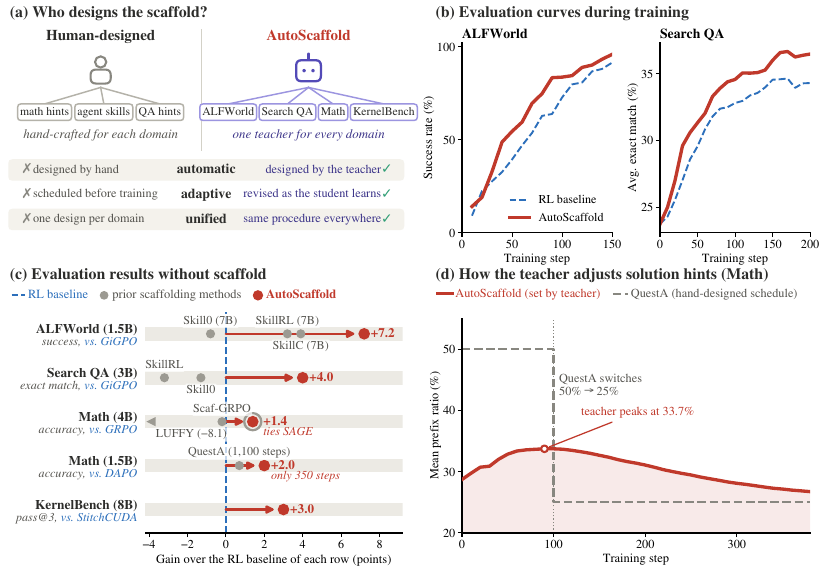}
\caption{\textbf{AutoScaffold: a teacher LLM designs and adapts scaffolds during RL training.}
(a) Unlike hand-designed scaffolding, AutoScaffold is automatic, adaptive, and unified.
(b) Evaluation curves during training (success rate on ALFWorld, exact match on search-based QA), compared with GiGPO.
(c) Gains over the RL baseline of each experiment, evaluated without scaffolds: success rate on ALFWorld, average exact match on search-based QA, average accuracy on math, and KernelBench correctness pass@3 averaged over Levels 1 to 3. Gray dots are prior scaffolding methods. All methods use the same student size as AutoScaffold, except Skill0, SkillRL and SkillC in ALFWorld using 7B models.
(d) The teacher sets each problem's solution-prefix ratio as instance-level scaffold: it first reveals more of the reference solutions and then withdraws support, whereas QuestA switches from 50\% to 25\% at a fixed step.}

\label{fig:teaser}
\end{figure}

Existing scaffolding methods, however, fall short in three respects. First, most are \emph{not automatic} and do not draw on the student's training experience: their scaffolds are prepared before training, as hand-designed hint tiers, partial solutions, or predefined skills~\citep{zhang2026scafgrposcaffoldedgrouprelative,li2026questaexpandingreasoningcapacity,lu2026skill0incontextagenticreinforcement}. A few methods derive support from the student's training experience, distilling trajectories into skills or generating guidance from failed attempts~\citep{xia2026skillrlevolvingagentsrecursive,he2026reskillreconcilingskillcreation}, but they do so through fixed pipelines. Second, they are \emph{not fully adaptive}. As the student learns, its abilities change: the failures a scaffold was designed for may disappear, and new failures appear that it does not address, so a scaffold that helped early in training may no longer help the student learn. Yet support either stays fixed or changes through predefined procedures: SAGE tunes hint strength with fixed rollout-based rules~\citep{liao2026selfhintinglanguagemodelsenhance}, Skill0 withdraws skills through a predefined staged curriculum~\citep{lu2026skill0incontextagenticreinforcement}, and SkillRL revises skills from a prescribed sample of failures without measuring their effect on the current student~\citep{xia2026skillrlevolvingagentsrecursive}. Third, they are \emph{not unified}: each is built around one task structure, such as hints or solution prefixes for mathematical reasoning or skill libraries for agentic tasks. A mathematical solution prefix, for example, does not provide the action sequencing or object-interaction strategies needed for an agentic task, so each new domain requires a new design. Deciding what support a domain needs, and how it should change over training, is therefore left to human researchers, who design scaffolds from their understanding of the domain and the model's failures, often refining them across repeated training runs.

These limitations motivate the central question of our work:
\par\nobreak\smallskip\nobreak
\begingroup
\setlength{\fboxsep}{6pt}
\noindent\fcolorbox{blue}{blue!4}{%
\parbox{\dimexpr\linewidth-2\fboxsep-2\fboxrule\relax}{%
\centering\itshape How can an LLM teacher turn a student's training experience into scaffolds, designing and adapting them throughout RL post-training, across domains, without human intervention?}}
\par
\endgroup
\smallskip

\noindent The difficulty is that the teacher's ability to solve a task does not by itself tell it how to help a weaker, evolving student learn. Making scaffolding \emph{automatic}, \emph{adaptive}, and \emph{unified} (Figure~\ref{fig:teaser}a) poses three challenges. 
\begin{itemize}
    \item \textbf{Diagnosing the student's bottlenecks.} Aggregate rewards reveal whether the student succeeds, but not why it fails, while a few trajectories may not represent a whole task class; the teacher must decide which evidence to inspect and which failure patterns warrant intervention. 

    \item \textbf{Choosing the right scaffold.} The teacher must turn each diagnosis into a scaffold that suits the domain and targets either a weakness shared by a task class or a single hard task. The scaffold must also be of the right strength, because if it is too weak, every attempt may fail, and if it is too strong, every attempt may succeed. In particular, under binary outcome rewards, both cases yield identical rewards within a rollout group and no reward-based policy-gradient signal. A scaffold that is too strong can also let the student succeed without learning to solve the task on its own. 
    \item \textbf{Validating and adapting scaffolds.} A scaffold tailored to a few observed failures may not help the student solve other tasks in the same class, and a scaffold that helps at the current point can become unnecessary as the student improves. The teacher must assess proposed changes and decide when to revise or withdraw scaffolds, with the goal of improving the student's performance without scaffolds.
\end{itemize}

Together, these challenges make obtaining effective scaffolds non-trivial, even with access to a strong teacher. They motivate a carefully designed framework that connects diagnosis, scaffold construction, and validation throughout training, so that the scaffold remains matched to the student's evolving needs.

\begin{figure}[t]
\centering
\includegraphics[width=\figw]{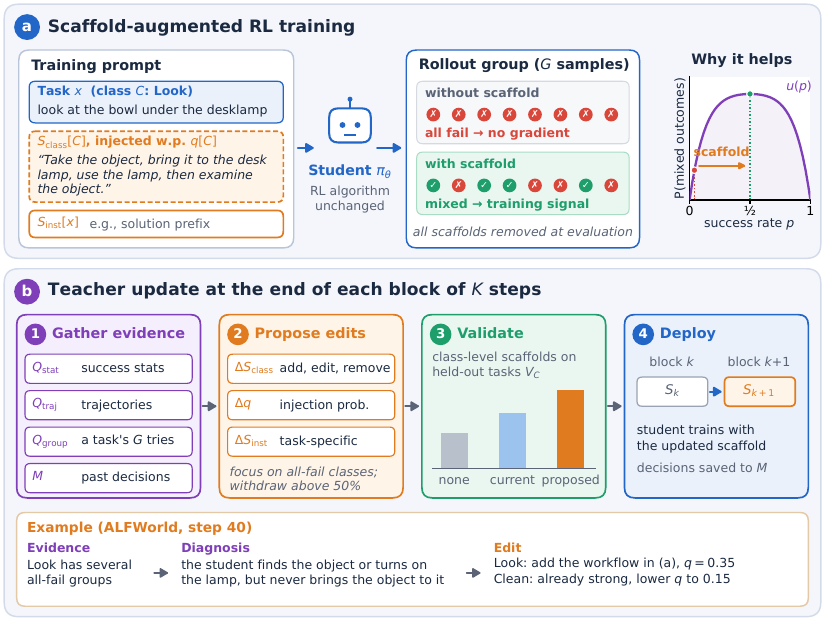}
\caption{\textbf{Overview of \textsc{AutoScaffold}.}
\textbf{(a)} A class-level scaffold, injected with probability $q[C]$, and an instance-level scaffold are added to the student's training prompt. On a hard task this turns all-failure groups into mixed groups that produce training signal; the RL algorithm is unchanged, and all scaffolds are removed at evaluation.
\textbf{(b)} At the end of each block of $K$ steps, a frozen teacher gathers evidence from the training log, proposes edits, validates class-level scaffolds on held-out tasks, and deploys the updated scaffold for the next block. Bottom: an actual update on ALFWorld.}
\label{fig:overview}

\end{figure}

In this paper, we introduce \textsc{AutoScaffold}, a unified AI-for-AI framework in which a frozen teacher LLM observes an ongoing RL training run and automatically turns the student's training experience into scaffolds injected into its training prompts (Figure~\ref{fig:hero}). The approach draws on the educational principle of scaffolding: temporary, learner-contingent support that enables a learner to accomplish tasks beyond their current independent ability, and that is withdrawn as competence develops~\citep{wood1976role,vygotsky1978mind}. AutoScaffold \emph{automatically} identifies the student's bottlenecks and proposes new scaffolds. Every $K$ training steps the teacher investigates the student's experience, using query tools to choose among training statistics, individual trajectories, groups of attempts on the same task, and its own past decisions (Figure~\ref{fig:overview}). To choose the right scaffold, it diagnoses why the student fails and writes two levels of scaffold: class-level strategies shared across a task class, and instance-level hints for tasks the student cannot yet solve, which play complementary roles in our analysis (Appendix~\ref{sec:theory}). Further, AutoScaffold also \emph{adaptively} changes the content and strength of the scaffold based on the student's current capability. The teacher revises the scaffolds every $K$ steps, and before deployment each class-level edit is validated on held-out tasks against no scaffold and the current scaffold, and the teacher uses these results to decide which edits to keep. It also revises how often class-level scaffolds are injected, and training with the updated scaffolds produces new experience for the next update. Finally, AutoScaffold \emph{unifies} the design of the scaffold mechanism across a wide spectrum of application domains, requiring minimal domain-specific input such as the task description and the task classes at the beginning of the training. 

We evaluate AutoScaffold on four domains with distinct task structures and RL algorithms: ALFWorld (embodied household tasks, GiGPO), search-based question answering (GiGPO), mathematical reasoning (GRPO and DAPO), and KernelBench (GPU kernel generation, rubric-based GRPO). Under scaffold-free evaluation, AutoScaffold improves over the corresponding RL baselines in every domain (Figure~\ref{fig:teaser}b,c), by 7.2 points in success rate on ALFWorld, 4.0 points in average exact match on search-based QA, up to 2.0 points in average accuracy on mathematical reasoning, and 3.0 points in correctness on KernelBench, and is competitive with or surpasses methods that rely on human-designed scaffolding. On KernelBench, it also raises the fraction of tasks with a correct kernel that runs faster than the PyTorch reference by up to 5.0 points (Table~\ref{tab:cuda-main}). On mathematical reasoning, it surpasses QuestA with approximately one-third as many student training steps. Ablations show the value of interactive queries and scaffold validation, and analyses of scaffold evolution show that, given only the general principles of scaffolding, the teacher decides on its own what scaffold to give, to which tasks, and how much, and revises these decisions as the student learns (Figure~\ref{fig:teaser}d).

Our main contributions are as follows:
\begin{itemize}
    \item \textbf{Automatic, adaptive scaffolding across domains.} We introduce \textsc{AutoScaffold}, a unified AI-for-AI framework that turns a student's training experience into evolving scaffolds through a frozen teacher LLM. The same procedure operates across domains with minimal domain-specific information and no human intervention during training.
    \item \textbf{A design that makes the teacher's scaffolds effective.} We design a framework that consists of a number of key components, such as adaptive scaffolding based on interactive queries and decision memory, controlled injection, and held-out validation of class-level revisions. Ablation studies show that each of these components is critical for the entire system; for example, removing scaffold validation reduces ALFWorld success by 14.1 points.
    \item \textbf{Principles for effective scaffolding.} Under binary rewards, we characterize conditions under which class- and instance-level scaffolds strengthen group-relative learning signals, and show how overly strong scaffolds can eliminate these signals (Appendix~\ref{sec:theory}). The analysis motivates combining strategies for shared weaknesses in a task class with hints for persistently unsolved tasks, and adapting support as the student learns.
    \item \textbf{Stronger performance without scaffolds.} Across embodied tasks, search-based QA, mathematical reasoning, and GPU kernel generation, \textsc{AutoScaffold} improves over the corresponding RL baselines under scaffold-free evaluation and is competitive with or surpasses scaffolding methods that rely heavily on human design, without hand-written scaffolds or hand-designed schedules.
\end{itemize}

\section{Related Work}
\label{sec:related}
\newcommand{\cmark}{\textcolor{teal}{\ding{51}}}
\newcommand{\xmark}{\textcolor{red!75!black}{\ding{55}}}
\newcommand{\pmark}{\textcolor{orange!85!black}{$\circ$}}

\paragraph{AI for automating model development.}
LLMs increasingly improve AI systems themselves, optimizing harnesses~\citep{lee2026metaharnessendtoendoptimizationmodel}, writing GPU kernels~\citep{ouyang2025kernelbenchllmswriteefficient,zhang2025cudaforgeagentframeworkhardware}, and conducting parts of research~\citep{lu2024aiscientistfullyautomated,yamada2025aiscientistv2workshoplevelautomated,karpathy2026autoresearch}. For model training, LLMs have designed rewards~\citep{ma2024eurekahumanlevelrewarddesign}, generated environments~\citep{chen2026traineetrainerllmdesignedtraining}, and tuned hyperparameters~\citep{liu2025largelanguagemodelagent}, and PostTrainBench~\citep{rank2026posttrainbenchllmagentsautomate} evaluates end-to-end automated post-training. Closest to our setting, LLMZero~\citep{fang2026llmzerodiscoveringadaptivetraining} diagnoses training signals at each checkpoint and searches over hyperparameter transitions for GRPO, and EvoTrainer~\citep{chen2026evotrainercoevolvingllmpolicies} co-evolves the policy with a training-side harness by diagnosing rollouts and backtesting interventions. These methods operate on the training configuration or harness; AutoScaffold instead intervenes within a run at the level of task classes and instances, turning the student's rollouts into guidance attached to its training prompts and removed at evaluation.

\paragraph{Scaffolding in RL for LLMs.}
To address tasks on which nearly every rollout fails, known as the learning cliff~\citep{zhang2026scafgrposcaffoldedgrouprelative}, prior work supplies task-specific hints, such as tiered hints in Scaf-GRPO~\citep{zhang2026scafgrposcaffoldedgrouprelative}, partial solutions in QuestA~\citep{li2026questaexpandingreasoningcapacity} and POPE~\citep{qu2026pope}, stepwise hints in StepHint~\citep{zhang2025stephint}, and self-generated hints in SAGE~\citep{liao2026selfhintinglanguagemodelsenhance}, or reusable skills for agentic tasks~\citep{xia2026skillrlevolvingagentsrecursive,lu2026skill0incontextagenticreinforcement,lin2026skillclearningautonomousskill,he2026reskillreconcilingskillcreation,shen2026slim}. In most methods, the form of guidance is designed by hand for one domain, and any adaptation follows predefined rules. The closest methods are SAGE, which adjusts hint strength with rollout-based rules; SkillRL, whose teacher revises skills from a fixed sample of failures without measuring their effect; and SLIM~\citep{shen2026slim}, which retains, retires, or adds skills by leave-one-skill-out validation but keeps some skills at inference. In contrast, AutoScaffold lets a frozen teacher choose which evidence to inspect, applies one procedure across domains, including math and GPU kernel generation, and validates class-level proposals on the current student before deployment. Table~\ref{tab:compare} positions AutoScaffold relative to both lines of work.

\begin{table}[t]
\caption{\textbf{Positioning of AutoScaffold.} \emph{Stage}: whether the method changes how a model behaves at inference time or how it is trained. \emph{What it changes}: the object that is designed or edited. \emph{Granularity}: whether guidance is shared by a class of tasks, specific to one task, or both. \emph{Automatic}: an AI system decides the changes, without hand-designed content or schedules (\pmark: content is generated by a model, but its use follows hand-designed rules). \emph{Adapts to student progress}: whether guidance or training settings are revised in response to the student's current performance and observed failures (\pmark: changes follow predefined rules or schedules). \emph{No extra runs}: changes are decided and tested within a single training run, without training separate runs or branches (\pmark: branches are trained from checkpoints). \emph{Scaffold-free eval.}: the learner is evaluated without the guidance it received during training (\pmark: some guidance is kept). \emph{Domains}: the task types on which the method is evaluated (A = agentic, S = search QA, M = math, K = GPU kernels). --: not applicable.}
\label{tab:compare}
\centering
\small
\setlength{\tabcolsep}{3.5pt}
\resizebox{\linewidth}{!}{
\begin{tabular}{lllcccccl}
\toprule
\textbf{Method} & \textbf{Stage} & \makecell[l]{\textbf{What it}\\\textbf{changes}} & \textbf{Granularity} & \textbf{Automatic} & \makecell{\textbf{Adapts to}\\\textbf{student progress}} & \makecell{\textbf{No extra}\\\textbf{runs}} & \makecell{\textbf{Scaffold-free}\\\textbf{eval.}} & \textbf{Domains} \\
\midrule
\multicolumn{9}{l}{\textit{AI for automating model development}} \\
AHE~\citep{lin2026agenticharnessengineeringobservabilitydriven} & inference & agent harness & -- & \cmark & -- & -- & -- & coding \\
autoresearch~\citep{karpathy2026autoresearch} & training & training code & -- & \cmark & \xmark & \xmark & -- & pretraining \\
Eureka~\citep{ma2024eurekahumanlevelrewarddesign} & training & reward function & -- & \cmark & \xmark & \xmark & -- & robotics \\
Trainee-to-Trainer~\citep{chen2026traineetrainerllmdesignedtraining} & training & environment & -- & \cmark & \cmark & \cmark & -- & multi-agent \\
LLMZero~\citep{fang2026llmzerodiscoveringadaptivetraining} & training & hyperparameters & -- & \cmark & \cmark & \pmark & -- & sci.\ QA, chem., music \\
EvoTrainer~\citep{chen2026evotrainercoevolvingllmpolicies} & training & recipe, diagnostics & -- & \cmark & \cmark & \xmark & -- & M, code, SWE \\
\midrule
\multicolumn{9}{l}{\textit{Scaffolding in RL for LLMs}} \\
Scaf-GRPO~\citep{zhang2026scafgrposcaffoldedgrouprelative} & training & hints & instance & \xmark & \pmark & \cmark & \cmark & M \\
QuestA~\citep{li2026questaexpandingreasoningcapacity} & training & partial solutions & instance & \xmark & \xmark & \cmark & \cmark & M \\
SAGE~\citep{liao2026selfhintinglanguagemodelsenhance} & training & hints & instance & \pmark & \pmark & \cmark & \cmark & M \\
SkillRL~\citep{xia2026skillrlevolvingagentsrecursive} & training & skills & class & \cmark & \cmark & \cmark & \xmark & A, S \\
Skill0~\citep{lu2026skill0incontextagenticreinforcement} & training & skills & class & \xmark & \pmark & \cmark & \cmark & A, S \\
SLIM~\citep{shen2026slim} & training & skills & class & \cmark & \cmark & \cmark & \xmark & A, S \\
\midrule
\textsc{AutoScaffold} (ours) & training & \makecell[l]{text scaffolds:\\skills, hints, partial solutions} & class + instance & \cmark & \cmark & \cmark & \cmark & A, S, M, K \\
\bottomrule
\end{tabular}}
\end{table}

\section{AutoScaffold}
\label{sec:method}

As mentioned in the introduction, an ideal scaffolding framework should be automatic, adaptive, and unified. Each requirement places demands on AutoScaffold. Being \emph{automatic} means that no human designs or reviews the scaffolds, so the teacher must find out on its own why the student fails, write the scaffolds itself, and check its own proposals before using them. Being \emph{adaptive} means that the scaffold must follow the current student, so it is revised throughout training, its effect is tracked across updates, and support is withdrawn once the student no longer needs it. Being \emph{unified} means that the same procedure must apply to any domain, so it relies only on minimal domain-specific information, allows different forms of textual scaffold across domains, leaves the RL algorithm unchanged, and guides the teacher with principles from our analysis rather than domain-specific rules. Figure~\ref{fig:overview} gives an overview of AutoScaffold. We describe how the student is trained with scaffolds (Section~\ref{sec:method:training}) and how a frozen teacher LLM designs them every $K$ training steps (Section~\ref{sec:method:generation}), noting which requirement each design serves.

\subsection{Scaffold-Augmented RL Training}
\label{sec:method:training}

This subsection explains why scaffolds help group-relative learning and why they must adapt to the student, defines scaffolds at two levels, and describes how they are injected into training prompts.

\paragraph{Why scaffolds help group-relative RL and why they must adapt.}
Group-relative methods such as GRPO~\citep{shao2024deepseekmathpushinglimitsmathematical} sample $G$ rollouts for a task $x$ and standardize each rollout's reward by the mean and standard deviation of its group. Under binary rewards, a group that contains both successes and failures yields advantages of roughly unit scale, whereas a group in which all rollouts fail, or all succeed, yields zero advantage and no gradient. Whether the model learns from a given task therefore depends on whether its rollout group contains mixed outcomes~\citep{liao2026selfhintinglanguagemodelsenhance}. If each rollout succeeds independently with probability $p$, then a mixed-outcome group occurs with probability
\begin{equation}
\label{eq:gate}
u(p) = 1 - (1-p)^G - p^G,
\end{equation}
which increases as $p$ approaches $\tfrac{1}{2}$ from either side and is approximately $Gp$ when $p$ is small (Proposition~\ref{prop:gate}). Tasks with $Gp \ll 1$ thus rarely contribute to training.\footnote{With mean-centered advantages that are not standardized~\citep{liu2025understanding}, the size of each update also scales with the within-group reward variance, whose expectation $\frac{G-1}{G}p(1-p)$ peaks at the same point (Proposition~\ref{prop:energy}). The analysis applies with minor modifications to bounded rewards such as rubric scores (Appendix~\ref{sec:theory}).} On a task the student rarely solves, a scaffold makes some rollouts succeed, so the group contains both outcomes and the task can generate training signal. Which scaffold achieves this, however, depends on the current student~\citep{liao2026selfhintinglanguagemodelsenhance}. As training proceeds, the student's abilities change, so the failures a scaffold was written for may disappear, and new failures appear that it does not address. Therefore, a scaffold that helped earlier may no longer help the student learn. Our ablation study further illustrates the limitations of fixed guidance: using Skill0's published skill bank as a fixed scaffold yields an ALFWorld success rate 15.3 points below AutoScaffold and 8.1 points below the RL baseline (Section~\ref{sec:ablation}).

\paragraph{Two levels of scaffold: class-level strategies and instance-level hints.}
A student may fail for two reasons: it may lack a strategy that a whole class of tasks requires, or a particular task may remain too hard for it even with such a strategy. Following the principle that support should match the capability that the learner currently lacks~\citep{wood1976role}, AutoScaffold provides scaffolds at two corresponding levels. To define the first level, we group training tasks into classes using the categories that each training dataset already provides, such as the six task types of ALFWorld or the training sources of search-based QA, in the same way that SkillRL and Skill0 organize skills by task category~\citep{xia2026skillrlevolvingagentsrecursive,lu2026skill0incontextagenticreinforcement}; when a dataset has no such categories, all its tasks form a single class. We write $C(x)$ for the class of task $x$, and Section~\ref{sec:exp:setup} lists the classes of each domain.
\begin{itemize}
    \item \textbf{Class-level scaffolds} $S_{\text{class}}[C]$ capture strategies shared by the tasks in class $C$, such as workflows (e.g., the order of steps that a type of household task requires), mistakes to avoid (e.g., moving an object again after it has been placed), and checks before finishing (e.g., confirming that the final answer is the quantity the problem asks for). Like the category-specific skills of SkillRL and Skill0, they transfer across tasks of the same class. By raising the success rate of a class, they increase the learning signal that the class provides, as long as they do not push its success rate far above one half (Proposition~\ref{prop:class}).
    \item \textbf{Instance-level scaffolds} $S_{\text{inst}}[x]$ give support specific to a single task $x$, such as a hint about its key step, an intermediate result, or part of its reference solution. One example is the solution prefix used by QuestA~\citep{li2026questaexpandingreasoningcapacity}, which reveals the first $r_x$ fraction of the reference solution; our math experiments use this form. Such scaffolds mainly help tasks that the student cannot solve even with a class-level strategy: they let these tasks produce successful rollouts and thus contribute to training at all (Proposition~\ref{prop:instance}).
\end{itemize}
The two levels are complementary: class-level scaffolds move tasks that the student can already partially solve toward a success rate near one half, where they give the most learning signal, while instance-level scaffolds bring tasks with zero success into training (Appendix~\ref{sec:theory}).

\paragraph{Injecting scaffolds into the student's training prompts.}
To apply to any domain and any group-relative method, every scaffold is plain text placed in the prompt, and the RL algorithm is left unchanged. A class-level scaffold is not shown to every rollout group of its class. Instead, each class $C$ has an injection probability $q[C]$, the fraction of its rollout groups that receive $S_{\text{class}}[C]$, so support for a class can be reduced gradually by lowering $q[C]$. The complete scaffold is thus $S = (S_{\text{class}}, q, S_{\text{inst}})$. For each rollout group of task $x$, we draw $z_x \sim \mathrm{Bernoulli}(q[C(x)])$ and build the prompt:
\begin{equation}
\label{eq:augment}
\tilde{x} = \mathrm{augment}\bigl(x,\ S^{(z_x)}_{\text{class}}[C(x)],\ S_{\text{inst}}[x]\bigr),
\end{equation}
where $S^{(1)}_{\text{class}}[C] = S_{\text{class}}[C]$, $S^{(0)}_{\text{class}}[C]$ is empty, and $S_{\text{inst}}[x]$ is empty for tasks without an instance-level scaffold. All $G$ rollouts in the group share the prompt $\tilde{x}$, so their rewards are compared under the same scaffold. Taking GRPO as an example, the student is trained to maximize
\begin{equation}
\label{eq:scaffold-grpo}
\mathcal{J}(\theta) = \mathbb{E}\Biggl[\frac{1}{G}\sum_{i=1}^{G}\frac{1}{|\tau_i|}\sum_{t=1}^{|\tau_i|}\min\Bigl(\rho_{i,t}A_i,\ \mathrm{clip}\bigl(\rho_{i,t},\,1-\varepsilon_{\text{clip}},\,1+\varepsilon_{\text{clip}}\bigr)A_i\Bigr)\Biggr],
\end{equation}
where the expectation is over scaffolded tasks $\tilde{x}$ built by Eq.~(\ref{eq:augment}) and rollouts $\{\tau_i\}_{i=1}^{G} \sim \pi_{\theta_{\text{old}}}(\cdot \mid \tilde{x})$, and
\begin{equation}
\label{eq:scaffold-ratio-adv}
\rho_{i,t} = \frac{\pi_\theta(\tau_{i,t} \mid \tilde{x}, \tau_{i,<t})}{\pi_{\theta_{\text{old}}}(\tau_{i,t} \mid \tilde{x}, \tau_{i,<t})}, \qquad
A_i = \frac{R(x,\tau_i) - \mathrm{mean}\bigl(\{R(x,\tau_j)\}_{j=1}^{G}\bigr)}{\mathrm{std}\bigl(\{R(x,\tau_j)\}_{j=1}^{G}\bigr) + \epsilon}
\end{equation}
are the importance ratio and the group-standardized advantage, $\varepsilon_{\text{clip}}$ is the clipping range, and $\epsilon \ge 0$ is a small constant for numerical stability; the KL penalty is omitted for brevity. Compared with standard GRPO, only the prompt changes: rollouts are sampled and their likelihoods computed conditioned on the scaffolded prompt $\tilde{x}$, while each rollout is scored by the original task reward $R(x,\tau_i)$. The scaffold is part of the prompt, so its tokens receive no gradient. Other group-relative methods used in our experiments, such as DAPO and GiGPO, are modified in the same way. Because all scaffolds are removed at evaluation, the student must also practice without help during training, so that it learns to solve tasks on its own rather than depend on the scaffold; these unscaffolded rollouts also let the teacher see when a scaffold is no longer needed. We therefore cap $q[C] \le q_{\max} < 1$, so each group has probability at least $1 - q_{\max}$ of being trained without the class-level scaffold, following privileged information dropout~\citep{kamienny2020privilegedinformationdropoutreinforcement}; in our ablation, injecting the class-level scaffold into every task ($q_{\max}=1$) lowers success below the RL baseline (Section~\ref{sec:ablation}). Each rollout is logged with its task, class, reward, and whether the class-level scaffold was injected, which lets the teacher compare the student's success with and without it. Algorithm~\ref{alg:training} summarizes the training procedure.

\begin{algorithm}[t]
\caption{Scaffold-augmented RL training}
\label{alg:training}
\small
\begin{algorithmic}[1]
\Require student $\pi_\theta$; training tasks $\mathcal{D}$ with classes $C(\cdot)$; group size $G$; initial scaffold $S^{0}$; update interval $K$; training steps $N$
\State Hold out $V_C$ from each class $C$ and set $\mathcal{D} \gets \mathcal{D} \setminus \bigcup_C V_C$ \Comment{held-out tasks for validation (Section~\ref{sec:method:generation})}
\State $S = (S_{\text{class}}, q, S_{\text{inst}}) \gets S^{0}$ \Comment{class-level scaffolds, injection probabilities, instance-level scaffolds}
\State $M \gets \varnothing$ \Comment{teacher's memory of its past diagnoses and decisions}
\State $\mathcal{L} \gets \varnothing$ \Comment{training log since the last scaffold update}
\For{$t = 1, \dots, N$}
    \State $\theta_{\text{old}} \gets \theta$; \ sample a batch of tasks $\mathcal{B} \subset \mathcal{D}$
    \For{each task $x \in \mathcal{B}$} \Comment{one rollout group per task}
        \State $z_x \sim \mathrm{Bernoulli}\bigl(q[C(x)]\bigr)$ \Comment{$z_x = 1$: inject $S_{\text{class}}[C(x)]$}
        \State $\tilde{x} \gets \mathrm{augment}\bigl(x,\ S^{(z_x)}_{\text{class}}[C(x)],\ S_{\text{inst}}[x]\bigr)$ \Comment{Eq.~(\ref{eq:augment})}
        \State Sample $\{\tau_i\}_{i=1}^{G} \sim \pi_{\theta_{\text{old}}}(\cdot \mid \tilde{x})$ and score each with the task reward $R(x, \tau_i)$
        \State Append $\bigl(x, C(x), z_x, \{\tau_i, R(x,\tau_i)\}_{i=1}^{G}\bigr)$ to $\mathcal{L}$
    \EndFor
    \State Update $\theta$ by maximizing $\mathcal{J}(\theta)$ on the rollout groups of $\mathcal{B}$ \Comment{Eq.~(\ref{eq:scaffold-grpo})}
    \If{$t \bmod K = 0$ \textbf{and} $t < N$} \Comment{end of block $k = t/K$}
        \State $(S, M) \gets \textsc{UpdateScaffold}(k, \pi_{\theta}, S, \mathcal{L}, M, \{V_C\})$ \Comment{Algorithm~\ref{alg:teacher}}
        \State $\mathcal{L} \gets \varnothing$
    \EndIf
\EndFor
\State \Return $\pi_\theta$ \Comment{evaluated without any scaffold}
\end{algorithmic}
\end{algorithm}

\subsection{Automatic Scaffold Design and Adaptation}
\label{sec:method:generation}

AutoScaffold updates the scaffold every $K$ RL training steps. We call each such period a \emph{block}: the scaffold stays fixed within a block, and at the end of block $k$ a frozen teacher LLM updates it for block $k+1$ (Figure~\ref{fig:overview}(b)). Training starts with no class-level scaffold, so every class-level scaffold the student sees is written by the teacher. The teacher makes all scaffold decisions; its instructions contain no domain-specific rules and differ across domains only in a short description of the task domain and the list of classes. The teacher may see reference answers of training tasks but never sees evaluation tasks. Each update takes four steps: gathering evidence, proposing edits, validating class-level scaffolds, and deploying the update. Algorithm~\ref{alg:teacher} summarizes one update.

\begin{algorithm}[t]
\caption{\textsc{UpdateScaffold}: one teacher update at the end of block $k$}
\label{alg:teacher}
\small
\begin{algorithmic}[1]
\Require block $k$; student $\pi_{\theta_k}$; scaffold $S$; block log $\mathcal{L}$; memory $M$; held-out tasks $\{V_C\}$; teacher $T$; query budget $B$; cap $q_{\max}$
\State $E \gets \{(v^{\text{none}}_C, v^{\text{cur}}_C)\}_C$: success rates of $\pi_{\theta_k}$ on each $V_C$ without and with $S_{\text{class}}[C]$; \ $b \gets 0$ \Comment{gathering evidence}
\While{$T$ requests more evidence \textbf{and} $b < B$} \Comment{$b$: number of queries so far}
    \State $E \gets E \cup \{$response to one query chosen by $T$: $Q_{\text{stat}}$, $Q_{\text{traj}}$, or $Q_{\text{group}}$ on $\mathcal{L}$, or a read of $M\}$; \ $b \gets b + 1$
\EndWhile
\State $(D, \Delta S) \gets T(E, S)$, with $\Delta S = (\Delta S_{\text{class}}, \Delta q, \Delta S_{\text{inst}})$ and $q[C] \in [0, q_{\max}]$ \Comment{proposing edits; $D$: diagnosis}
\For{each class $C$ for which $\Delta S_{\text{class}}$ proposes new text $S'_C$} \Comment{validating class-level scaffolds}
    \State $v^{\text{new}}_C \gets$ success rate of $\pi_{\theta_k}$ on $V_C$ with $S'_C$
    \State $T$ revises $\Delta S_{\text{class}}[C]$ based on $v^{\text{none}}_C$, $v^{\text{cur}}_C$, and $v^{\text{new}}_C$
\EndFor
\State Apply $\Delta S$ to $S$ \Comment{deploying the update}
\State $M \gets M \cup \{(k, D, \Delta S)\}$ \Comment{record the diagnosis and decisions}
\State \Return $S, M$
\end{algorithmic}
\end{algorithm}

\paragraph{Gathering evidence.}
Since no human diagnoses the student, the teacher must find out on its own why the student fails, and which evidence answers this is not known in advance. A low success rate in a class may come from one shared mistake or from several unrelated ones; telling them apart requires reading trajectories, comparing failed attempts with successful ones on the same task, and checking whether a pattern seen in one trajectory recurs across the class. Unlike prior methods, which give the teacher a fixed sample of failures, AutoScaffold lets the teacher decide what to read. The teacher accesses the training log of the current block through three query tools, each returning a view of the log at a different level of detail:
\begin{itemize}
    \item \textbf{Class statistics.} $Q_{\text{stat}}$ returns each class's success rates with and without its class-level scaffold, and the numbers of tasks whose rollouts all failed or all succeeded.
    \item \textbf{Individual trajectories.} $Q_{\text{traj}}$ returns trajectories filtered by class and outcome.
    \item \textbf{Rollout groups.} $Q_{\text{group}}$ returns all $G$ attempts on one task, showing how successful and failed attempts on the same task differ.
\end{itemize}
The teacher gathers evidence in a multi-turn loop. At each turn, it either calls one tool, whose response is appended to its context, or stops and proposes edits. Starting from initial evidence, such as each class's success rate on its held-out tasks $V_C$ (described below) with and without the current scaffold, it chooses which trajectories or groups to read next based on what it has already seen; each response is kept within its context, and the teacher must stop after $B$ calls. The log of one block, however, shows only how the student behaves now, not how it responded to the teacher's earlier edits. The teacher therefore keeps a memory $M$ of its past diagnoses and decisions, and can query $M$ for its latest three decisions to judge whether those edits helped. For example, at step 40 on ALFWorld, the teacher first calls $Q_{\text{stat}}$, finds that Pick2, Look, and Heat are the weakest classes, then calls $Q_{\text{traj}}$ to read four all-failure trajectories from each of them, and finally reads its last three decisions from memory (Figure~\ref{fig:case}, top). Replacing the query tools and memory with a fixed sample of failed trajectories, as in SkillRL, lowers ALFWorld success by 8.6 points (Section~\ref{sec:ablation}).

\paragraph{Proposing edits.}
From this evidence, the teacher diagnoses why the student fails in each class and proposes an edit $\Delta S = (\Delta S_{\text{class}}, \Delta q, \Delta S_{\text{inst}})$, which covers three decisions:
\begin{itemize}
    \item \textbf{Class-level strategies.} The teacher adds, revises, or removes $S_{\text{class}}[C]$ to address a weakness shared across class $C$. It writes these scaffolds as general strategies rather than answers to specific tasks. For example, after observing that the student in ALFWorld's \emph{Pick2} class (two-object placement tasks) often places only one object, the teacher adds a strategy for finding and placing the second object (Section~\ref{sec:analysis}).
    \item \textbf{Injection frequency.} The teacher sets $q[C] \in [0, q_{\max}]$ for each class, for example lowering $q[C]$ as the student succeeds in class $C$ without the scaffold.
    \item \textbf{Instance-level hints.} The teacher writes or revises $S_{\text{inst}}[x]$, or adjusts the fraction $r_x$ of a verified reference solution shown to the student, giving more support to tasks the student keeps failing and less to tasks it can already solve. As a simple illustration, a hint for $x^4-3x^2+1=0$ could suggest substituting $t=x^2$, leaving the student to solve the resulting quadratic and recover $x$.
\end{itemize}
Instead of domain-specific rules, the teacher's instructions state two principles from the analysis in Section~\ref{sec:method:training}, which hold in every domain. First, focus on classes with low success rates or many all-failure groups, where scaffolds add the most learning signal (Propositions~\ref{prop:class} and~\ref{prop:instance}). Second, reduce support as the student improves; for class-level scaffolds, start withdrawing once success without the scaffold exceeds 50\%, a heuristic threshold motivated by our analysis (Propositions~\ref{prop:gate} and~\ref{prop:class}). The teacher may also leave the scaffold unchanged. Appendix~\ref{app:case_studies} gives examples of its diagnoses and edits.

\paragraph{Validating class-level scaffolds.}
A class-level scaffold is inferred from a few trajectories, so it may fix the failures the teacher has read without helping other tasks in the same class. Because it is shown to the whole class for an entire block, a harmful scaffold affects many training steps. AutoScaffold therefore checks each proposed class-level scaffold on the current student before deploying it:
\begin{itemize}
    \item \textbf{Held-out tasks.} Before training, we reserve a set of tasks $V_C$ from the training data of each class and exclude them from training. At the start of each update, the current student is evaluated on $V_C$ without a class-level scaffold and with the current one, and these success rates are part of the initial evidence.
    \item \textbf{Three conditions.} The success rates on $V_C$ without a class-level scaffold and with the current scaffold come from the evaluation at the start of the update. For each proposed scaffold, we additionally run the current student $\pi_{\theta_k}$, the checkpoint at the end of block $k$, on $V_C$ with the proposed scaffold, using $n$ rollouts per task and matched seeds.
    \item \textbf{Comparisons.} The teacher receives the three success rates ($v^{\text{none}}_C$, $v^{\text{cur}}_C$, and $v^{\text{new}}_C$ in Algorithm~\ref{alg:teacher}). Comparing the proposed scaffold with the current one shows whether the edit improves on existing guidance; comparing either with no scaffold shows whether the student still benefits from support in this class.
    \item \textbf{Decision.} Using these results and its diagnosis, the teacher adopts the proposed scaffold, keeps the current one, combines the two, or removes the scaffold of the class.
\end{itemize}
A higher success rate under the scaffold is not by itself the goal: the teacher prefers scaffolds that help the student succeed while leaving meaningful work for it, and how often the student sees a scaffold is set separately by $q[C]$. Without validation, ALFWorld success drops from 93.9\% to 79.8\%, below the RL baseline (Section~\ref{sec:ablation}). Instance-level scaffolds are not validated, since in our experiments they are prefixes of verified reference solutions rather than text written by the teacher.

\paragraph{Deploying the update.}
The accepted edits are applied to $S$ automatically in a fixed format and form the scaffold for block $k+1$. The teacher's diagnosis and decisions are added to $M$, and training under the new scaffold produces the evidence for the next update. Repeating this process lets the scaffold follow the student: at representative ALFWorld updates, continuing training with the revised scaffold outperforms both keeping the previous scaffold and using none (Section~\ref{sec:analysis}).

\section{Experiments}
\begingroup
\setlength{\intextsep}{6pt plus 1pt minus 1pt}
\setlength{\textfloatsep}{8pt plus 1pt minus 1pt}
\setlength{\floatsep}{6pt plus 1pt minus 1pt}

Our experiments test whether AutoScaffold delivers the three properties argued for in Section~\ref{sec:intro}. \textbf{Automatic}: without hand-written scaffolds or hand-designed schedules, does it improve the student's performance without scaffolds over RL baselines, and is it competitive with methods built on human-designed scaffolds? \textbf{Unified}: does the same procedure work across domains with different task structures and RL algorithms? Section~\ref{sec:exp:main} answers both. \textbf{Adaptive}: does revising scaffolds during training matter, and how does the teacher design and adjust them as the student learns? Sections~\ref{sec:ablation} and~\ref{sec:analysis} answer this.

\subsection{Experimental Setup}
\label{sec:exp:setup}

\noindent\textbf{Domains and tasks.} We evaluate four domains with distinct task structures: \textbf{ALFWorld}, with six household task types (Pick, Look, Clean, Heat, Cool, Pick2) requiring planning and object interaction~\citep{shridhar2021alfworldaligningtextembodied}; \textbf{search-based QA}, covering single-hop (NQ, TriviaQA, PopQA) and multi-hop questions (HotpotQA, 2Wiki, MuSiQue, Bamboogle)~\citep{kwiatkowski-etal-2019-natural,joshi2017triviaqalargescaledistantly,mallen2023trustlanguagemodelsinvestigating,yang2018hotpotqadatasetdiverseexplainable,ho2020constructingmultihopqadataset,trivedi2022musiquemultihopquestionssinglehop,press2023measuringnarrowingcompositionalitygap}; \textbf{KernelBench}, with GPU kernel tasks ranging from single operators to full networks~\citep{ouyang2025kernelbenchllmswriteefficient}; and \textbf{math reasoning}, covering eight competition-level benchmarks~\citep{aime2024,aime2025,hmmt2025feb,brumo2025,amc2023,lightman2023letsverifystepstep,lewkowycz2022solvingquantitativereasoningproblems,he2024olympiadbenchchallengingbenchmarkpromoting}. Task classes follow each training dataset: the six task types in ALFWorld, the two training sources (NQ and HotpotQA) in search-based QA, six kernel types in KernelBench, and a single class for math (Table~\ref{tab:classes}). Appendix~\ref{app:domains} provides details.

\begin{table}[!htb]
\caption{Task classes used for class-level scaffolds in each domain. Classes follow the categories provided by each training dataset; math has no such categories and forms a single class.}
\label{tab:classes}
\centering
\small
\begin{tabular}{lll}
\toprule
\textbf{Domain} & \textbf{Source of classes} & \textbf{Classes} \\
\midrule
ALFWorld & task type & Pick, Look, Clean, Heat, Cool, Pick2 \\
Search-based QA & training source & NQ, HotpotQA \\
KernelBench & kernel type & Conv, Elementwise, Loss, Matmul, Norm/Softmax, Reduce \\
Math reasoning & none & a single class \\
\bottomrule
\end{tabular}
\end{table}

\noindent\textbf{Baselines.} We compare with three types of baselines: \textbf{(i) training-free methods}, which prompt the base model without any training; \textbf{(ii) training without scaffolds}, which train the student with the same task reward but no scaffold; and \textbf{(iii) training with scaffolds}, which add human-designed scaffolds, such as skills, hints, or partial solutions, to the student's prompts during training.

For \textit{ALFWorld}, we compare with (i) ReAct~\citep{yao2023reactsynergizingreasoningacting} and Reflexion~\citep{shinn2023reflexionlanguageagentsverbal}; (ii) PPO~\citep{schulman2017proximalpolicyoptimizationalgorithms}, RLOO~\citep{ahmadian2024basicsrevisitingreinforcestyle}, GRPO~\citep{shao2024deepseekmathpushinglimitsmathematical}, and GiGPO~\citep{feng2025gigpo}; and (iii) SkillRL~\citep{xia2026skillrlevolvingagentsrecursive}, Skill0~\citep{lu2026skill0incontextagenticreinforcement}, and SkillC~\citep{lin2026skillclearningautonomousskill}, which are reported only with Qwen2.5-7B-Instruct and serve as references.
For \textit{search-based QA}, we compare with (ii) R1-Instruct, Search-R1~\citep{jin2025searchr1}, ZeroSearch~\citep{sun2026zerosearchincentivizesearchcapability}, StepSearch~\citep{wang2025stepsearchignitingllmssearch}, and GiGPO; and (iii) SkillRL and Skill0, at both the same scale as our student and 7B.
For \textit{KernelBench}, we compare with (ii) StitchCUDA~\citep{li2026stitchcudaautomatedmultiagentsendtoend}, the rubric-based RL method whose training setting we follow.
For \textit{math reasoning}, we follow two settings. With Qwen3-4B-Instruct, we compare with (ii) SFT and GRPO, and (iii) LUFFY~\citep{yan2025learningreasonoffpolicyguidance}, Scaf-GRPO~\citep{zhang2026scafgrposcaffoldedgrouprelative}, and SAGE~\citep{liao2026selfhintinglanguagemodelsenhance}. With Nemotron-1.5B, we compare with (ii) DAPO~\citep{yu2025dapoopensourcellmreinforcement} and (iii) QuestA~\citep{li2026questaexpandingreasoningcapacity}.
Baseline results are taken from the original papers or from the work whose training settings we follow, as noted in each table. Among them, SkillRL keeps its skills in the prompt at evaluation, whereas AutoScaffold is always evaluated without scaffolds.

\noindent\textbf{Implementation.} In all domains, the teacher is GPT-5.5 with default reasoning effort and uses the same instructions apart from the domain description and class list. Scaffold updates occur every $K=10$ steps with $q_{\max}=0.5$, and each update allows at most $B=15$ teacher queries. The teacher writes class-level scaffolds in all domains, while instance-level scaffolds are used only for math. Students and RL algorithms follow prior work in each domain: GiGPO's settings for ALFWorld and QA, with Qwen2.5-1.5B-Instruct and Qwen2.5-3B-Instruct; StitchCUDA's rubric-based RL for KernelBench, with a Qwen3-8B student cold-started as in Dr.~Kernel~\citep{liu2026drkernelreinforcementlearning} and 600 training tasks; and SAGE (Qwen3-4B-Instruct with GRPO) and QuestA (Nemotron-1.5B with DAPO) for math, using their official training data. Experiments use 4--8 H200 GPUs, and results are averaged over 3 independent runs. The teacher's API cost is a few dollars per run (Appendix~\ref{app:cost}). \textbf{We evaluate AutoScaffold without scaffolds}, so the results measure the student's own capability.

\subsection{Main Results}
\label{sec:exp:main}

Across the four domains, the same framework, with teacher instructions that differ only in the domain description and class list, improves over the corresponding RL baseline under four RL algorithms (GiGPO, rubric-based GRPO, GRPO, and DAPO) and five students. In every domain where scaffolding methods exist, it also matches or surpasses them, although it starts from no class-level scaffold and uses no hand-written scaffold or hand-designed schedule.

\noindent\textbf{ALFWorld.} AutoScaffold reaches 93.9\% success, 7.2 points above GiGPO, and is best on five of the six task classes (Table~\ref{tab:alfworld-main}). The largest gains are on Look (+19.7), Cool (+16.2), and Pick2 (+15.3), the three classes where GiGPO is weakest and successful rollouts are rarest. The 1.5B student also exceeds SkillRL (89.9\%) and SkillC (90.6\%), which use the larger Qwen2.5-7B-Instruct, and SkillRL additionally keeps its skills at evaluation.

\noindent\textbf{Search-based QA.} AutoScaffold achieves 46.1\% average exact match, 4.0 points above GiGPO, and is best among 3B models on six of seven benchmarks (Table~\ref{tab:qa-main}). At the same scale, it exceeds SkillRL and Skill0 by 7.2 and 5.3 points, and it also surpasses 7B Skill0 (44.4\%). Although training uses only NQ and HotpotQA, the gain over GiGPO on the five out-of-domain benchmarks (+4.1 on average) is as large as on the two in-domain ones (+3.8), so the scaffolds improve skills that transfer beyond the training data.

\noindent\textbf{Mathematical reasoning.} With Qwen3-4B-Instruct, AutoScaffold reaches 70.1\% average accuracy, 1.4 points above GRPO, and is on par with SAGE (70.0\%), the strongest scaffolding method, without SAGE's hand-designed hint levels and scheduling rules (Table~\ref{tab:math-qwen3-4b}). Guidance does not help by default: LUFFY, which trains on off-policy reference traces, falls 5.2 points below the base model. With Nemotron-1.5B, AutoScaffold reaches 62.4\%, 2.0 points above DAPO and 1.3 points above QuestA, using 350 training steps compared with QuestA's 1,100, roughly one-third (Table~\ref{tab:math-main}). The 1.5B student also surpasses DeepSeek-R1-Distill-32B (58.1\%).

\noindent\textbf{GPU kernel generation.} AutoScaffold improves all six KernelBench metrics over StitchCUDA (Table~\ref{tab:cuda-main}), raising correctness from 59\% to 63\% on Level 1 and from 50\% to 54\% on Level 2, and Level 2 Fast$_1$ from 20\% to 25\%. Gains on the full-network tasks of Level 3 are smaller (+1 point on both metrics). Since this domain uses rubric-based rather than binary rewards, these results also show that AutoScaffold is not limited to the binary-reward setting of our analysis.

\begin{table}[!htb]
\caption{Success rates (\%) on ALFWorld with Qwen2.5-1.5B-Instruct. Overall for AutoScaffold is the unweighted mean over the six task types. Prompting and RL baseline results are from \citet{feng2025gigpo}; ours are mean $\pm$ standard deviation over 3 seeds. Gray rows are from \citet{lin2026skillclearningautonomousskill} and use the larger Qwen2.5-7B-Instruct (reference only). \textbf{Bold}: best among 1.5B models.}
\label{tab:alfworld-main}
\centering
\small
\setlength{\tabcolsep}{4.5pt}
\renewcommand{\arraystretch}{0.9}
\begin{tabular}{lccccccc}
\toprule
\textbf{Method} & Pick & Look & Clean & Heat & Cool & Pick2 & Overall \\
\midrule
Qwen2.5-1.5B-Instruct & 5.9 & 5.5 & 3.3 & 9.7 & 4.2 & 0.0 & 4.1 \\
\midrule
\multicolumn{8}{c}{\textit{Training-free methods}} \\
ReAct & 17.4 & 20.5 & 15.7 & 6.2 & 7.7 & 2.0 & 12.8 \\
Reflexion & 35.3 & 22.2 & 21.7 & 13.6 & 19.4 & 3.7 & 21.8 \\
\midrule
\multicolumn{8}{c}{\textit{RL baselines without scaffold}} \\
PPO & 64.8 & 40.5 & 57.1 & 60.6 & 46.4 & 47.4 & 54.4 \\
RLOO & 88.3 & 52.8 & 71.0 & 62.8 & 66.4 & 56.9 & 69.7 \\
GRPO & 85.3 & 53.7 & 84.5 & 78.2 & 59.7 & 53.5 & 72.8 \\
GiGPO & 94.4 & 67.5 & 94.8 & \textbf{94.4} & 79.8 & 76.4 & 86.7 \\
\midrule
\multicolumn{8}{c}{\textit{RL baselines with scaffold (Qwen2.5-7B-Instruct, reference only)}} \\
\gc{SkillRL} & \gc{97.9} & \gc{71.4} & \gc{90.0} & \gc{90.0} & \gc{95.5} & \gc{87.5} & \gc{89.9} \\
\gc{Skill0} & \gc{100.0} & \gc{84.6} & \gc{91.1} & \gc{84.2} & \gc{88.9} & \gc{55.5} & \gc{85.9} \\
\gc{SkillC} & \gc{88.5} & \gc{78.6} & \gc{91.2} & \gc{94.7} & \gc{100.0} & \gc{88.9} & \gc{90.6} \\
\midrule
\textsc{AutoScaffold} (ours) & \textbf{100.0}{\scriptsize$\pm$0.0} & \textbf{87.2}{\scriptsize$\pm$4.4} & \textbf{95.1}{\scriptsize$\pm$2.1} & 93.8{\scriptsize$\pm$0.0} & \textbf{96.0}{\scriptsize$\pm$0.0} & \textbf{91.7}{\scriptsize$\pm$4.2} & \textbf{93.9}{\scriptsize$\pm$0.3} \\
\bottomrule
\end{tabular}
\end{table}

\begin{table}[!htb]
\caption{Exact match (\%) on search-based QA with Qwen2.5-3B-Instruct, trained on NQ and HotpotQA ($\dagger$: in-domain; $\star$: out-of-domain). RL baseline results are from \citet{feng2025gigpo}; SkillRL and Skill0 are from their original papers~\citep{xia2026skillrlevolvingagentsrecursive,lu2026skill0incontextagenticreinforcement}; ours are mean $\pm$ standard deviation over 3 seeds. Gray rows use the larger Qwen2.5-7B-Instruct (reference only). \textbf{Bold}: best among 3B models.}
\label{tab:qa-main}
\centering
\footnotesize
\setlength{\tabcolsep}{2.9pt}
\renewcommand{\arraystretch}{0.9}
\begin{tabular}{l ccc cccc c}
\toprule
\multirow{2}{*}{\textbf{Method}} & \multicolumn{3}{c}{\textbf{Single-Hop QA}} & \multicolumn{4}{c}{\textbf{Multi-Hop QA}} & \multirow{2}{*}{\textbf{Avg.}} \\
\cmidrule(lr){2-4} \cmidrule(lr){5-8}
 & NQ$^\dagger$ & TriviaQA$^\star$ & PopQA$^\star$ & HotpotQA$^\dagger$ & 2Wiki$^\star$ & MuSiQue$^\star$ & Bamboogle$^\star$ & \\
\midrule
\multicolumn{9}{c}{\textit{RL baselines without scaffold}} \\
R1-Instruct & 27.0 & 53.7 & 19.9 & 23.7 & 29.2 & 7.2 & 29.3 & 27.1 \\
Search-R1 & 34.1 & 54.5 & 37.8 & 32.4 & 31.9 & 10.3 & 26.4 & 32.5 \\
ZeroSearch & 41.4 & 57.4 & 44.8 & 27.4 & 30.0 & 9.8 & 11.1 & 31.7 \\
StepSearch & -- & -- & -- & 34.5 & 32.0 & \textbf{17.4} & 34.4 & -- \\
GiGPO & 42.0 & 59.5 & 42.4 & 36.9 & 37.0 & 12.6 & 64.1 & 42.1 \\
\midrule
\multicolumn{9}{c}{\textit{RL baselines with scaffold}} \\
SkillRL & 38.6 & 57.6 & 40.3 & 33.6 & 31.1 & 13.3 & 58.1 & 38.9 \\
Skill0 & 39.8 & 57.5 & 42.3 & 35.1 & 33.7 & 13.3 & 63.7 & 40.8 \\
\midrule
\multicolumn{9}{c}{\textit{RL baselines with scaffold (Qwen2.5-7B-Instruct, reference only)}} \\
\gc{SkillRL} & \gc{45.9} & \gc{63.3} & \gc{45.9} & \gc{43.2} & \gc{40.3} & \gc{20.2} & \gc{73.8} & \gc{47.1} \\
\gc{Skill0} & \gc{42.7} & \gc{61.1} & \gc{45.3} & \gc{40.0} & \gc{38.3} & \gc{16.4} & \gc{66.9} & \gc{44.4} \\
\midrule
\textsc{AutoScaffold} (ours) & \textbf{46.3}{\scriptsize$\pm$1.3} & \textbf{62.3}{\scriptsize$\pm$2.0} & \textbf{49.6}{\scriptsize$\pm$1.2} & \textbf{40.2}{\scriptsize$\pm$0.4} & \textbf{39.7}{\scriptsize$\pm$0.6} & 16.6{\scriptsize$\pm$0.5} & \textbf{67.7}{\scriptsize$\pm$0.5} & \textbf{46.1}{\scriptsize$\pm$0.6} \\
\bottomrule
\end{tabular}
\end{table}

\begin{table}[!htb]
\caption{Accuracy (\%) on math benchmarks with Qwen3-4B-Instruct. RL baseline results are from \citet{liao2026selfhintinglanguagemodelsenhance}; ours are mean $\pm$ standard deviation over 3 seeds. Avg.: mean over the six benchmarks. \textbf{Bold} / \underline{underline}: best / second best.}
\label{tab:math-qwen3-4b}
\centering
\footnotesize
\setlength{\tabcolsep}{4pt}
\renewcommand{\arraystretch}{0.9}
\begin{tabular}{lccccccc}
\toprule
\textbf{Method} & AIME24 & AIME25 & AMC23 & MATH-500 & Minerva & Olympiad & Avg. \\
\midrule
Qwen3-4B-Instruct & 52.1 & 43.1 & 92.2 & 93.6 & 46.1 & 67.7 & 65.8 \\
\midrule
\multicolumn{8}{c}{\textit{RL baselines without scaffold}} \\
SFT & 14.4 & 22.1 & 55.2 & 78.9 & 38.1 & 40.2 & 41.5 \\
GRPO & 55.8 & 45.0 & \textbf{95.0} & \textbf{96.0} & \textbf{50.1} & 70.4 & 68.7 \\
\midrule
\multicolumn{8}{c}{\textit{RL baselines with scaffold}} \\
LUFFY & 42.3 & 36.0 & 86.4 & 91.1 & 48.2 & 59.4 & 60.6 \\
Scaf-GRPO & \textbf{59.8} & 45.2 & 92.2 & 95.1 & 48.9 & 69.8 & 68.5 \\
SAGE-\textsc{light} & 59.2 & 47.1 & 92.2 & 95.1 & \underline{49.5} & 70.5 & 68.9 \\
SAGE & 58.1 & \textbf{52.1} & \underline{94.2} & 95.4 & 49.2 & \textbf{71.2} & \underline{70.0} \\
\midrule
\textsc{AutoScaffold} (ours) & \underline{59.3}{\scriptsize$\pm$0.4} & \underline{51.4}{\scriptsize$\pm$0.4} & 93.9{\scriptsize$\pm$0.5} & \underline{95.5}{\scriptsize$\pm$0.1} & 49.4{\scriptsize$\pm$0.2} & \underline{71.0}{\scriptsize$\pm$0.5} & \textbf{70.1}{\scriptsize$\pm$0.3} \\
\bottomrule
\end{tabular}
\end{table}

\begin{table}[!htb]
\caption{Accuracy (\%) on math benchmarks with Nemotron-1.5B. DAPO and QuestA results are from \citet{li2026questaexpandingreasoningcapacity}. Gray: larger models (reference only). \textbf{Bold}: best among small models.}
\label{tab:math-main}
\centering
\footnotesize
\setlength{\tabcolsep}{4.5pt}
\renewcommand{\arraystretch}{0.9}
\begin{tabular}{lcccccc}
\toprule
\textbf{Method} & AIME24 & AIME25 & HMMT Feb 25 & OlympiadBench & BRUMO25 & Avg. \\
\midrule
\multicolumn{7}{c}{\textit{Training-free base models}} \\
DeepSeek-R1-Distill-1.5B & 28.7 & 22.3 & 12.0 & 52.4 & 31.8 & 29.4 \\
Qwen3-1.7B & 48.3 & 36.8 & 22.2 & 56.1 & 44.1 & 41.5 \\
Nemotron-1.5B & 61.8 & 49.5 & 31.6 & 64.6 & 58.2 & 53.1 \\
\midrule
\multicolumn{7}{c}{\textit{Larger models (reference only)}} \\
\gc{DeepSeek-R1-Distill-32B} & \gc{72.6} & \gc{51.8} & \gc{33.0} & \gc{65.0} & \gc{68.0} & \gc{58.1} \\
\gc{Qwen3-8B} & \gc{76.0} & \gc{67.3} & \gc{44.8} & \gc{68.6} & \gc{68.3} & \gc{65.0} \\
\midrule
\multicolumn{7}{c}{\textit{RL baselines without scaffold}} \\
DAPO (2K steps) & 69.5 & 59.8 & 38.9 & 68.1 & 66.0 & 60.4 \\
\midrule
\multicolumn{7}{c}{\textit{RL baselines with scaffold}} \\
QuestA (1.1K steps) & 69.3 & \textbf{60.0} & 37.9 & 69.7 & 68.3 & 61.1 \\
\midrule
\textsc{AutoScaffold} (ours, 350 steps) & \textbf{72.6} & 58.1 & \textbf{42.1} & \textbf{70.2} & \textbf{69.2} & \textbf{62.4} \\
\bottomrule
\end{tabular}
\end{table}

\begin{table}[!htb]
\caption{Pass@3 of Correctness and Fast$_1$ on the KernelBench test set. Fast$_1$: the generated kernel is correct and faster than the PyTorch reference~\citep{ouyang2025kernelbenchllmswriteefficient}. \textbf{Bold} / \underline{underline}: best / second best.}
\label{tab:cuda-main}
\centering
\small
\setlength{\tabcolsep}{5pt}
\renewcommand{\arraystretch}{0.9}
\begin{tabular}{lcccccc}
\toprule
\multirow{2}{*}{\textbf{Method}} & \multicolumn{2}{c}{Level 1} & \multicolumn{2}{c}{Level 2} & \multicolumn{2}{c}{Level 3} \\
\cmidrule(lr){2-3} \cmidrule(lr){4-5} \cmidrule(lr){6-7}
 & Correctness & Fast$_1$ & Correctness & Fast$_1$ & Correctness & Fast$_1$ \\
\midrule
Qwen3-8B-Cold-Start & 51 & 14 & 43 & 16 & \underline{32} & 8 \\
StitchCUDA & \underline{59} & \underline{17} & \underline{50} & \underline{20} & \underline{32} & \underline{13} \\
\midrule
\textsc{AutoScaffold} (ours) & \textbf{63} & \textbf{19} & \textbf{54} & \textbf{25} & \textbf{33} & \textbf{14} \\
\bottomrule
\end{tabular}
\end{table}

\subsection{Ablation Study}
\label{sec:ablation}

We evaluate the contribution of AutoScaffold's main design choices through ablations on ALFWorld with Qwen2.5-1.5B-Instruct (Table~\ref{tab:ablation}). Each of the four main designs matters: removing the query tools and memory, removing validation, fixing the scaffold, or always injecting it drops the student below the RL baseline, so the gains come from how scaffolds are designed and used rather than from extra guidance alone.

\begin{table}[!htb]
\caption{Ablations on ALFWorld (Qwen2.5-1.5B-Instruct), success rate (\%) evaluated without scaffold. Static: Skill0's published ALFWorld skill bank used as a fixed class-level scaffold. $\Delta$: gap to full \textsc{AutoScaffold}.}
\label{tab:ablation}
\centering
\small
\setlength{\tabcolsep}{3.5pt}
\renewcommand{\arraystretch}{0.9}
\begin{tabular}{l cc cc cccc}
\toprule
 & \multirow{2}{*}{\textsc{AutoScaffold}} & \multirow{2}{*}{GiGPO} & \multicolumn{2}{c}{\textit{Teacher decision}} & \multicolumn{4}{c}{\textit{Scaffold adaptation \& injection}} \\
\cmidrule(lr){4-5} \cmidrule(lr){6-9}
 & & & w/o tools \& memory & w/o validation & Static & $K{=}5$ & $K{=}20$ & $q_{\max}{=}1$ \\
\midrule
Success & \textbf{93.9} & 86.7 & 85.3 & 79.8 & 78.6 & 90.1 & 90.4 & 86.0 \\
$\Delta$ & -- & $-$7.2 & $-$8.6 & $-$14.1 & $-$15.3 & $-$3.8 & $-$3.5 & $-$7.9 \\
\bottomrule
\end{tabular}
\end{table}

\noindent\textbf{Query tools and memory.} Replacing the query tools and memory with a fixed sample of failed trajectories and a single teacher call, as in SkillRL~\citep{xia2026skillrlevolvingagentsrecursive}, lowers success to 85.3\%. With a fixed sample of failures only, the teacher cannot compare failures with successes, look further when the sample is inconclusive, or check how the student responded to its earlier edits.

\noindent\textbf{Scaffold validation.} Deploying every proposal without validation causes the largest drop among teacher components, to 79.8\%, 6.9 points below GiGPO. Because a class-level scaffold is shown to a whole class for an entire block, an unhelpful edit affects many updates before the teacher can revise it.

\noindent\textbf{Static scaffolds.} Using Skill0's published ALFWorld skill bank~\citep{lu2026skill0incontextagenticreinforcement} as a fixed class-level scaffold lowers success to 78.6\%, 15.3 points below AutoScaffold and 8.1 points below GiGPO. The bank was not built for this student and cannot change as the student's failures change; Section~\ref{sec:analysis} shows how much the teacher's own scaffold changes over training.

\noindent\textbf{Update interval $K$.} Updating every 5 or 20 steps gives 90.1\% and 90.4\%, both above GiGPO but below the default $K=10$. Shorter blocks give the teacher less evidence per update, while longer blocks let the scaffold lag behind the student; $K=10$ also needs half as many teacher calls as $K=5$.

\noindent\textbf{Injection probability.} Injecting the class-level scaffold into every rollout group ($q_{\max}=1$) lowers success to 86.0\%, slightly below GiGPO, although the scaffold content is produced in the same way. The student then never practices without help, which it must do at evaluation (Section~\ref{sec:method:training}).

\subsection{Analysis}
\label{sec:analysis}

The ablations show that each design is needed. This section looks at what the teacher actually does with them. We first follow a single update in detail, then trace how the scaffold changes over a full run, test whether revisions help the student at the time they are made, compare the scaffolds the teacher writes in different domains, and finally follow how the teacher adjusts instance-level scaffolds in math.

\begin{figure}[t]
\centering
\includegraphics[width=\figw]{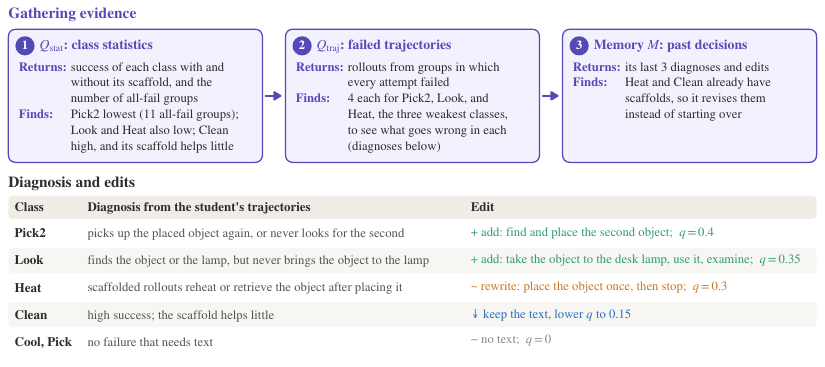}
\caption{\textbf{One teacher update on ALFWorld (step 40).} Top: the teacher queries tools and memory, then gathers evidence from class statistics, failed trajectories, and its own past decisions. It reads them and finds meaningful information for further diagnosis and edits. Bottom: it diagnoses a different failure in each class and responds with a matching edit: adding, rewriting, reducing, or omitting a scaffold.}
\label{fig:case}
\end{figure}

\noindent\textbf{How the teacher decides at one update (ALFWorld, step 40).} Figure~\ref{fig:case} follows this update. The teacher checks class statistics with $Q_{\text{stat}}$, inspects four all-failure trajectories per weak class (Pick2, Look, and Heat) using $Q_{\text{traj}}$, and reviews its last three decisions in memory $M$. Its decisions depend on the needs of each class.
\begin{itemize}
    \item \textbf{Pick2.} The student picks up the first object after placing it or never searches for the second. The teacher adds a strategy for finding and placing the second object.
    \item \textbf{Look.} The student finds the object or lamp but fails to bring the object to the lamp. The teacher specifies taking the object to the desk lamp, using the lamp, and then examining the object.
    \item \textbf{Heat.} Even with the existing scaffold, the student reheats or retrieves the object after placing it. The teacher rewrites the scaffold to emphasize placing the object once and stopping.
    \item \textbf{Clean.} The student already succeeds often and gains little from the scaffold, so the teacher keeps its text but lowers its injection probability.
    \item \textbf{Cool and Pick.} The teacher identifies no failure requiring a scaffold and provides no text.
\end{itemize}
Before deployment, each proposed class-level scaffold is also validated on $V_C$. Appendix~\ref{app:case:alfworld} gives another case, from a separate run, that records the full process of one update: how the teacher gathers evidence, what it diagnoses and proposes, and how validation decides which scaffolds are adopted.

\noindent\textbf{Why these scaffolds help the student learn.} These decisions show how the teacher turns observed failures into useful guidance. For Look tasks, the scaffold adds the missing step of bringing the object to the lamp. By helping some rollouts succeed, the scaffold can turn an all-failure group into a mixed group with a useful learning signal. The scaffolds describe strategies that apply across tasks in a class, while leaving the student to select objects and carry out the actions. The teacher also adjusts its guidance based on the student's response. It clarifies the Heat scaffold when the student repeats actions after placing the object and reduces support for Clean when the scaffold provides little additional benefit.

\begin{figure}[t]
\centering
\includegraphics[width=\figw]{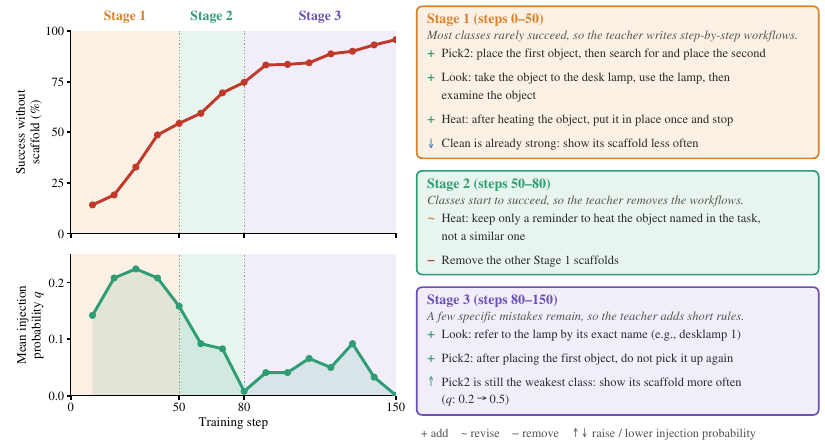}
\caption{\textbf{Scaffold changes over training on ALFWorld.} Left: scaffold-free success (top) and mean class-level injection probability (bottom), with training divided into three stages. Right: what the teacher changed in each stage (+ add, $\sim$ revise, $-$ remove, $\uparrow\downarrow$ raise or lower injection probability).}
\label{fig:scaffold_trend}
\end{figure}

\noindent\textbf{How the scaffold changes over a full run.} Across training, the teacher adjusts both the content of each class's scaffold and how often it is used. The ALFWorld run in Figure~\ref{fig:scaffold_trend} illustrates three stages.
\begin{itemize}
    \item \textbf{Stage~1 (steps 0--50).} When success is low across most classes, the teacher writes procedures for two-object, examine-in-light, and heating tasks. The mean injection probability rises, giving the student more frequent support.
    \item \textbf{Stage~2 (steps 50--80).} As success improves, the teacher removes the broad workflows and retains a targeted reminder to heat the specified object. The mean injection probability falls close to zero.
    \item \textbf{Stage~3 (steps 80--150).} The teacher addresses remaining failures with specific reminders, such as using the lamp's exact name in Look and leaving the first object where it was placed in Pick2. It raises the injection probability for the still-difficult Pick2 class.
\end{itemize}
The support thus evolves from broad procedures to targeted fixes, with changes guided by each class's training evidence. For comparison, Skill0 uses a withdrawal schedule fixed before training.

\begin{figure}[!htb]
\centering
\includegraphics[width=0.85\figw]{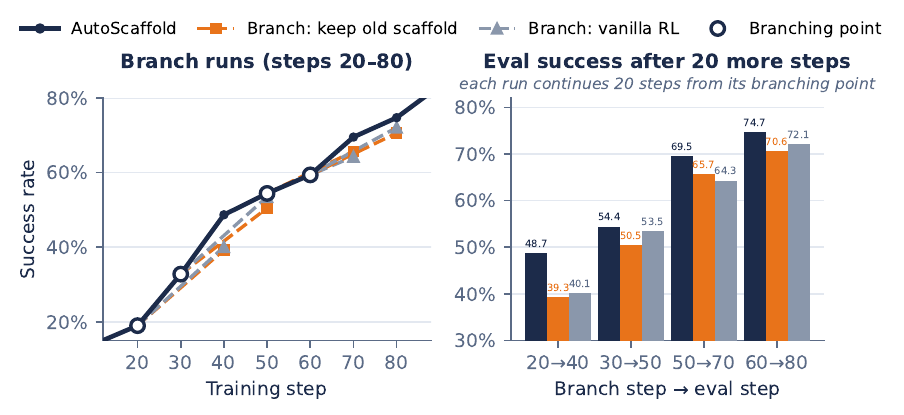}
\caption{From each branching point, we train 20 more steps with the updated scaffold (\textsc{AutoScaffold}, dark blue), with the previous scaffold kept fixed (orange), or with vanilla RL (gray). \textsc{AutoScaffold} gives the highest success at every branching point.}
\label{fig:counterfactual}
\end{figure}

\noindent\textbf{Do revisions help the student when they are made?} We measure the effect of scaffold revisions while holding the starting student checkpoint and number of subsequent training steps fixed. At representative ALFWorld checkpoints between steps 20 and 80, we create three branches and train each for 20 additional steps using the updated scaffold (AutoScaffold), the previous scaffold kept fixed, or no scaffold (vanilla RL). The updated scaffold yields the highest success at every tested checkpoint (Figure~\ref{fig:counterfactual}). Its advantage over the previous scaffold shows that the revisions improve learning over the following 20 steps. Its advantage over vanilla RL shows that support remains useful at these checkpoints.

\noindent\textbf{What scaffolds the teacher writes in different domains.} The teacher adapts scaffold content to the failures observed in each domain. In ALFWorld, its guidance describes action sequences for household tasks. In math, its class-level notes target recurring errors in otherwise sound reasoning, such as stopping after an intermediate result or boxing a quantity other than the one the problem asks for (Appendix~\ref{app:case:math}).

\begin{figure}[t]
\centering
\includegraphics[width=0.6\figw]{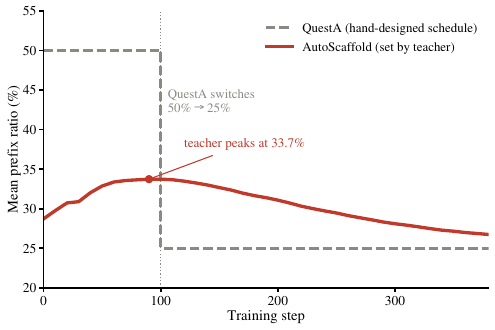}
\caption{\textbf{Solution-prefix ratios during math training (Nemotron-1.5B).} Mean ratio across all training problems. AutoScaffold (red) adjusts each problem's ratio using training evidence. QuestA (gray) switches from 50\% to 25\% at step 100.}
\label{fig:math_ratio}
\end{figure}

\noindent\textbf{How the teacher adjusts solution hints in math.} We also examine how the teacher adjusts instance-level hints during math training, where each hint reveals a fraction $r_x$ of the problem's reference solution (Figure~\ref{fig:math_ratio}). We initialize the hints with QuestA's data~\citep{li2026questaexpandingreasoningcapacity}, in which each problem has a 50\% or 25\% solution prefix, mix both sets from the start, and let the teacher adjust the prefix ratios alongside class-level guidance. During the first 100 steps, the teacher reveals more of the reference solutions, raising the mean prefix ratio from 28.7\% to 33.7\%. The mean ratio then declines to 26.7\% by the end of training; at step 100, for example, the teacher raises the prefixes of all-fail problems that still have room and lowers them for problems the student already solves (Appendix~\ref{app:case:math}).
Our instructions ask the teacher to reduce support for problems the student can already solve. The teacher determines the timing and amount of each adjustment from training evidence. QuestA follows a fixed schedule, using 50\% prefixes for the first 100 steps and 25\% thereafter. AutoScaffold begins withdrawing support near this scheduled switch and reaches a similar final mean ratio through adjustments to individual problems. Together with the ALFWorld results, this illustrates how the same framework adapts both class-level strategies and instance-level hints as the student learns.
\endgroup
\section{Conclusion}
We presented \textsc{AutoScaffold}, an AI-for-AI framework that uses a frozen teacher to turn a student's RL training experience into adaptive guidance. The teacher gathers evidence through query tools and memory, designs class-level strategies and instance-level hints, and validates class-level proposals before use. It adjusts both the guidance and the amount of support as the student learns. Across four domains, the same procedure improves performance over the corresponding RL baselines under scaffold-free evaluation and is competitive with or surpasses methods built on human-designed scaffolds. The gains include both correctness and execution efficiency on GPU kernel generation. Ablations and controlled comparisons support the value of evidence gathering, validation, and scaffold adaptation. Future work includes reducing the cost of teacher interventions and characterizing when learning with scaffolds improves the student's independent capabilities.

\section{\textbf{Disclaimer}}

This paper was prepared for informational purposes by
the Artificial Intelligence Research group of JPMorgan Chase \& Co. and its affiliates (``JP Morgan''),
and is not a product of the Research Department of JP Morgan.
JP Morgan makes no representation and warranty whatsoever and disclaims all liability,
for the completeness, accuracy or reliability of the information contained herein.
This document is not intended as investment research or investment advice, or a recommendation,
offer or solicitation for the purchase or sale of any security, financial instrument, financial product or service,
or to be used in any way for evaluating the merits of participating in any transaction,
and shall not constitute a solicitation under any jurisdiction or to any person,
if such solicitation under such jurisdiction or to such person would be unlawful.

\bibliography{references}

@inproceedings{feng2025gigpo,
  title={Group-in-Group Policy Optimization for {LLM} Agent Training},
  author={Feng, Lang and Xue, Zhenghai and Liu, Tingcong and An, Bo},
  booktitle={Advances in Neural Information Processing Systems},
  year={2025}
}

@article{jin2025searchr1,
  title={Search-{R1}: Training {LLM}s to Reason and Leverage Search Engines with Reinforcement Learning},
  author={Jin, Bowen and Zeng, Hansi and Yue, Zhenrui and Yoon, Jinsung and Arik, Sercan and Wang, Dong and Zamani, Hamed and Han, Jiawei},
  journal={arXiv preprint arXiv:2503.09516},
  year={2025}
}

@misc{shinn2023reflexionlanguageagentsverbal,
      title={Reflexion: Language Agents with Verbal Reinforcement Learning}, 
      author={Noah Shinn and Federico Cassano and Edward Berman and Ashwin Gopinath and Karthik Narasimhan and Shunyu Yao},
      year={2023},
      eprint={2303.11366},
      archivePrefix={arXiv},
      primaryClass={cs.AI},
      url={https://arxiv.org/abs/2303.11366}, 
}

@misc{lee2026metaharnessendtoendoptimizationmodel,
      title={Meta-Harness: End-to-End Optimization of Model Harnesses}, 
      author={Yoonho Lee and Roshen Nair and Qizheng Zhang and Kangwook Lee and Omar Khattab and Chelsea Finn},
      year={2026},
      eprint={2603.28052},
      archivePrefix={arXiv},
      primaryClass={cs.AI},
      url={https://arxiv.org/abs/2603.28052}, 
}

@misc{lin2026agenticharnessengineeringobservabilitydriven,
      title={Agentic Harness Engineering: Observability-Driven Automatic Evolution of Coding-Agent Harnesses}, 
      author={Jiahang Lin and Shichun Liu and Chengjun Pan and Lizhi Lin and Shihan Dou and Zhiheng Xi and Xuanjing Huang and Hang Yan and Zhenhua Han and Tao Gui and Yu-Gang Jiang},
      year={2026},
      eprint={2604.25850},
      archivePrefix={arXiv},
      primaryClass={cs.CL},
      url={https://arxiv.org/abs/2604.25850}, 
}

@misc{shao2024deepseekmathpushinglimitsmathematical,
      title={DeepSeekMath: Pushing the Limits of Mathematical Reasoning in Open Language Models}, 
      author={Zhihong Shao and Peiyi Wang and Qihao Zhu and Runxin Xu and Junxiao Song and Xiao Bi and Haowei Zhang and Mingchuan Zhang and Y. K. Li and Y. Wu and Daya Guo},
      year={2024},
      eprint={2402.03300},
      archivePrefix={arXiv},
      primaryClass={cs.CL},
      url={https://arxiv.org/abs/2402.03300}, 
}

@misc{yu2025dapoopensourcellmreinforcement,
      title={DAPO: An Open-Source LLM Reinforcement Learning System at Scale}, 
      author={Qiying Yu and Zheng Zhang and Ruofei Zhu and Yufeng Yuan and Xiaochen Zuo and Yu Yue and Weinan Dai and Tiantian Fan and Gaohong Liu and Lingjun Liu and Xin Liu and Haibin Lin and Zhiqi Lin and Bole Ma and Guangming Sheng and Yuxuan Tong and Chi Zhang and Mofan Zhang and Wang Zhang and Hang Zhu and Jinhua Zhu and Jiaze Chen and Jiangjie Chen and Chengyi Wang and Hongli Yu and Yuxuan Song and Xiangpeng Wei and Hao Zhou and Jingjing Liu and Wei-Ying Ma and Ya-Qin Zhang and Lin Yan and Mu Qiao and Yonghui Wu and Mingxuan Wang},
      year={2025},
      eprint={2503.14476},
      archivePrefix={arXiv},
      primaryClass={cs.LG},
      url={https://arxiv.org/abs/2503.14476}, 
}

@article{wood1976role,
  title={The role of tutoring in problem solving},
  author={Wood, David and Bruner, Jerome S and Ross, Gail},
  journal={Journal of child psychology and psychiatry},
  volume={17},
  number={2},
  pages={89--100},
  year={1976},
  publisher={Blackwell Publishing Ltd Oxford, UK}
}

@misc{vygotsky1978mind,
  title={Mind in society: The development of higher psychological processes},
  author={Vygotsky, Lev S},
  year={1978},
  publisher={Harvard university press}
}

@misc{liao2026selfhintinglanguagemodelsenhance,
      title={Self-Hinting Language Models Enhance Reinforcement Learning}, 
      author={Baohao Liao and Hanze Dong and Xinxing Xu and Christof Monz and Jiang Bian},
      year={2026},
      eprint={2602.03143},
      archivePrefix={arXiv},
      primaryClass={cs.LG},
      url={https://arxiv.org/abs/2602.03143}, 
}

@misc{chen2026evotrainercoevolvingllmpolicies,
      title={EvoTrainer: Co-Evolving LLM Policies and Training Harnesses for Autonomous Agentic Reinforcement Learning}, 
      author={Guhong Chen and Yingcheng Shi and Yongbin Li and Binhua Li and Xander Xu and Hu Wei and Shiwen Ni and Min Yang and Jieping Ye},
      year={2026},
      eprint={2606.03108},
      archivePrefix={arXiv},
      primaryClass={cs.AI},
      url={https://arxiv.org/abs/2606.03108}, 
}

@misc{kamienny2020privilegedinformationdropoutreinforcement,
      title={Privileged Information Dropout in Reinforcement Learning}, 
      author={Pierre-Alexandre Kamienny and Kai Arulkumaran and Feryal Behbahani and Wendelin Boehmer and Shimon Whiteson},
      year={2020},
      eprint={2005.09220},
      archivePrefix={arXiv},
      primaryClass={cs.LG},
      url={https://arxiv.org/abs/2005.09220}, 
}

@misc{yao2023reactsynergizingreasoningacting,
      title={ReAct: Synergizing Reasoning and Acting in Language Models}, 
      author={Shunyu Yao and Jeffrey Zhao and Dian Yu and Nan Du and Izhak Shafran and Karthik Narasimhan and Yuan Cao},
      year={2023},
      eprint={2210.03629},
      archivePrefix={arXiv},
      primaryClass={cs.CL},
      url={https://arxiv.org/abs/2210.03629}, 
}

@misc{xia2026skillrlevolvingagentsrecursive,
      title={SkillRL: Evolving Agents via Recursive Skill-Augmented Reinforcement Learning}, 
      author={Peng Xia and Jianwen Chen and Hanyang Wang and Jiaqi Liu and Kaide Zeng and Yu Wang and Siwei Han and Yiyang Zhou and Xujiang Zhao and Haifeng Chen and Zeyu Zheng and Cihang Xie and Huaxiu Yao},
      year={2026},
      eprint={2602.08234},
      archivePrefix={arXiv},
      primaryClass={cs.LG},
      url={https://arxiv.org/abs/2602.08234}, 
}

@misc{lu2026skill0incontextagenticreinforcement,
      title={SKILL0: In-Context Agentic Reinforcement Learning for Skill Internalization}, 
      author={Zhengxi Lu and Zhiyuan Yao and Jinyang Wu and Chengcheng Han and Qi Gu and Xunliang Cai and Weiming Lu and Jun Xiao and Yueting Zhuang and Yongliang Shen},
      year={2026},
      eprint={2604.02268},
      archivePrefix={arXiv},
      primaryClass={cs.LG},
      url={https://arxiv.org/abs/2604.02268}, 
}

@misc{lin2026skillclearningautonomousskill,
      title={SKILLC: Learning Autonomous Skill Internalization in LLM Agents via Contrastive Credit Assignment}, 
      author={Hongxiang Lin and Zhirui Kuai and Erpeng Xue and Lei Wang},
      year={2026},
      eprint={2605.27899},
      archivePrefix={arXiv},
      primaryClass={cs.AI},
      url={https://arxiv.org/abs/2605.27899}, 
}

@misc{yan2025learningreasonoffpolicyguidance,
      title={Learning to Reason under Off-Policy Guidance}, 
      author={Jianhao Yan and Yafu Li and Zican Hu and Zhi Wang and Ganqu Cui and Xiaoye Qu and Yu Cheng and Yue Zhang},
      year={2025},
      eprint={2504.14945},
      archivePrefix={arXiv},
      primaryClass={cs.LG},
      url={https://arxiv.org/abs/2504.14945}, 
}

@misc{zhang2026scafgrposcaffoldedgrouprelative,
      title={Scaf-GRPO: Scaffolded Group Relative Policy Optimization for Enhancing LLM Reasoning}, 
      author={Xichen Zhang and Sitong Wu and Yinghao Zhu and Haoru Tan and Shaozuo Yu and Ziyi He and Jiaya Jia},
      year={2026},
      eprint={2510.19807},
      archivePrefix={arXiv},
      primaryClass={cs.CL},
      url={https://arxiv.org/abs/2510.19807}, 
}

@misc{li2026questaexpandingreasoningcapacity,
      title={QuestA: Expanding Reasoning Capacity in LLMs via Question Augmentation}, 
      author={Jiazheng Li and Hongzhou Lin and Hong Lu and Kaiyue Wen and Zaiwen Yang and Jiaxuan Gao and Yi Wu and Jingzhao Zhang},
      year={2026},
      eprint={2507.13266},
      archivePrefix={arXiv},
      primaryClass={cs.CL},
      url={https://arxiv.org/abs/2507.13266}, 
}

@misc{schulman2017proximalpolicyoptimizationalgorithms,
      title={Proximal Policy Optimization Algorithms}, 
      author={John Schulman and Filip Wolski and Prafulla Dhariwal and Alec Radford and Oleg Klimov},
      year={2017},
      eprint={1707.06347},
      archivePrefix={arXiv},
      primaryClass={cs.LG},
      url={https://arxiv.org/abs/1707.06347}, 
}

@misc{ahmadian2024basicsrevisitingreinforcestyle,
      title={Back to Basics: Revisiting REINFORCE Style Optimization for Learning from Human Feedback in LLMs}, 
      author={Arash Ahmadian and Chris Cremer and Matthias Gallé and Marzieh Fadaee and Julia Kreutzer and Olivier Pietquin and Ahmet Üstün and Sara Hooker},
      year={2024},
      eprint={2402.14740},
      archivePrefix={arXiv},
      primaryClass={cs.LG},
      url={https://arxiv.org/abs/2402.14740}, 
}

@misc{sun2026zerosearchincentivizesearchcapability,
      title={ZeroSearch: Incentivize the Search Capability of LLMs without Searching}, 
      author={Hao Sun and Zile Qiao and Jiayan Guo and Xuanbo Fan and Yingyan Hou and Yong Jiang and Pengjun Xie and Yan Zhang and Fei Huang and Jingren Zhou},
      year={2026},
      eprint={2505.04588},
      archivePrefix={arXiv},
      primaryClass={cs.CL},
      url={https://arxiv.org/abs/2505.04588}, 
}

@misc{wang2025stepsearchignitingllmssearch,
      title={StepSearch: Igniting LLMs Search Ability via Step-Wise Proximal Policy Optimization}, 
      author={Ziliang Wang and Xuhui Zheng and Kang An and Cijun Ouyang and Jialu Cai and Yuhang Wang and Yichao Wu},
      year={2025},
      eprint={2505.15107},
      archivePrefix={arXiv},
      primaryClass={cs.CL},
      url={https://arxiv.org/abs/2505.15107}, 
}

@misc{li2026stitchcudaautomatedmultiagentsendtoend,
      title={StitchCUDA: An Automated Multi-Agents End-to-End GPU Programing Framework with Rubric-based Agentic Reinforcement Learning}, 
      author={Shiyang Li and Zijian Zhang and Winson Chen and Yuebo Luo and Mingyi Hong and Caiwen Ding},
      year={2026},
      eprint={2603.02637},
      archivePrefix={arXiv},
      primaryClass={cs.MA},
      url={https://arxiv.org/abs/2603.02637}, 
}

@misc{shridhar2021alfworldaligningtextembodied,
      title={ALFWorld: Aligning Text and Embodied Environments for Interactive Learning}, 
      author={Mohit Shridhar and Xingdi Yuan and Marc-Alexandre Côté and Yonatan Bisk and Adam Trischler and Matthew Hausknecht},
      year={2021},
      eprint={2010.03768},
      archivePrefix={arXiv},
      primaryClass={cs.CL},
      url={https://arxiv.org/abs/2010.03768}, 
}

@article{kwiatkowski-etal-2019-natural,
    title = "Natural Questions: A Benchmark for Question Answering Research",
    author = "Kwiatkowski, Tom  and
      Palomaki, Jennimaria  and
      Redfield, Olivia  and
      Collins, Michael  and
      Parikh, Ankur  and
      Alberti, Chris  and
      Epstein, Danielle  and
      Polosukhin, Illia  and
      Devlin, Jacob  and
      Lee, Kenton  and
      Toutanova, Kristina  and
      Jones, Llion  and
      Kelcey, Matthew  and
      Chang, Ming-Wei  and
      Dai, Andrew M.  and
      Uszkoreit, Jakob  and
      Le, Quoc  and
      Petrov, Slav",
    editor = "Lee, Lillian  and
      Johnson, Mark  and
      Roark, Brian  and
      Nenkova, Ani",
    journal = "Transactions of the Association for Computational Linguistics",
    volume = "7",
    year = "2019",
    address = "Cambridge, MA",
    publisher = "MIT Press",
    url = "https://aclanthology.org/Q19-1026/",
    doi = "10.1162/tacl_a_00276",
    pages = "452--466"
}

@misc{joshi2017triviaqalargescaledistantly,
      title={TriviaQA: A Large Scale Distantly Supervised Challenge Dataset for Reading Comprehension}, 
      author={Mandar Joshi and Eunsol Choi and Daniel S. Weld and Luke Zettlemoyer},
      year={2017},
      eprint={1705.03551},
      archivePrefix={arXiv},
      primaryClass={cs.CL},
      url={https://arxiv.org/abs/1705.03551}, 
}

@misc{mallen2023trustlanguagemodelsinvestigating,
      title={When Not to Trust Language Models: Investigating Effectiveness of Parametric and Non-Parametric Memories}, 
      author={Alex Mallen and Akari Asai and Victor Zhong and Rajarshi Das and Daniel Khashabi and Hannaneh Hajishirzi},
      year={2023},
      eprint={2212.10511},
      archivePrefix={arXiv},
      primaryClass={cs.CL},
      url={https://arxiv.org/abs/2212.10511}, 
}

@misc{yang2018hotpotqadatasetdiverseexplainable,
      title={HotpotQA: A Dataset for Diverse, Explainable Multi-hop Question Answering}, 
      author={Zhilin Yang and Peng Qi and Saizheng Zhang and Yoshua Bengio and William W. Cohen and Ruslan Salakhutdinov and Christopher D. Manning},
      year={2018},
      eprint={1809.09600},
      archivePrefix={arXiv},
      primaryClass={cs.CL},
      url={https://arxiv.org/abs/1809.09600}, 
}

@misc{ho2020constructingmultihopqadataset,
      title={Constructing A Multi-hop QA Dataset for Comprehensive Evaluation of Reasoning Steps}, 
      author={Xanh Ho and Anh-Khoa Duong Nguyen and Saku Sugawara and Akiko Aizawa},
      year={2020},
      eprint={2011.01060},
      archivePrefix={arXiv},
      primaryClass={cs.CL},
      url={https://arxiv.org/abs/2011.01060}, 
}

@misc{trivedi2022musiquemultihopquestionssinglehop,
      title={MuSiQue: Multihop Questions via Single-hop Question Composition}, 
      author={Harsh Trivedi and Niranjan Balasubramanian and Tushar Khot and Ashish Sabharwal},
      year={2022},
      eprint={2108.00573},
      archivePrefix={arXiv},
      primaryClass={cs.CL},
      url={https://arxiv.org/abs/2108.00573}, 
}

@misc{press2023measuringnarrowingcompositionalitygap,
      title={Measuring and Narrowing the Compositionality Gap in Language Models}, 
      author={Ofir Press and Muru Zhang and Sewon Min and Ludwig Schmidt and Noah A. Smith and Mike Lewis},
      year={2023},
      eprint={2210.03350},
      archivePrefix={arXiv},
      primaryClass={cs.CL},
      url={https://arxiv.org/abs/2210.03350}, 
}

@misc{ouyang2025kernelbenchllmswriteefficient,
      title={KernelBench: Can LLMs Write Efficient GPU Kernels?}, 
      author={Anne Ouyang and Simon Guo and Simran Arora and Alex L. Zhang and William Hu and Christopher Ré and Azalia Mirhoseini},
      year={2025},
      eprint={2502.10517},
      archivePrefix={arXiv},
      primaryClass={cs.LG},
      url={https://arxiv.org/abs/2502.10517}, 
}

@misc{lightman2023letsverifystepstep,
      title={Let's Verify Step by Step}, 
      author={Hunter Lightman and Vineet Kosaraju and Yura Burda and Harri Edwards and Bowen Baker and Teddy Lee and Jan Leike and John Schulman and Ilya Sutskever and Karl Cobbe},
      year={2023},
      eprint={2305.20050},
      archivePrefix={arXiv},
      primaryClass={cs.LG},
      url={https://arxiv.org/abs/2305.20050}, 
}

@misc{lewkowycz2022solvingquantitativereasoningproblems,
      title={Solving Quantitative Reasoning Problems with Language Models}, 
      author={Aitor Lewkowycz and Anders Andreassen and David Dohan and Ethan Dyer and Henryk Michalewski and Vinay Ramasesh and Ambrose Slone and Cem Anil and Imanol Schlag and Theo Gutman-Solo and Yuhuai Wu and Behnam Neyshabur and Guy Gur-Ari and Vedant Misra},
      year={2022},
      eprint={2206.14858},
      archivePrefix={arXiv},
      primaryClass={cs.CL},
      url={https://arxiv.org/abs/2206.14858}, 
}

@misc{he2024olympiadbenchchallengingbenchmarkpromoting,
      title={OlympiadBench: A Challenging Benchmark for Promoting AGI with Olympiad-Level Bilingual Multimodal Scientific Problems}, 
      author={Chaoqun He and Renjie Luo and Yuzhuo Bai and Shengding Hu and Zhen Leng Thai and Junhao Shen and Jinyi Hu and Xu Han and Yujie Huang and Yuxiang Zhang and Jie Liu and Lei Qi and Zhiyuan Liu and Maosong Sun},
      year={2024},
      eprint={2402.14008},
      archivePrefix={arXiv},
      primaryClass={cs.CL},
      url={https://arxiv.org/abs/2402.14008}, 
}

@misc{chen2026traineetrainerllmdesignedtraining,
      title={From Trainee to Trainer: LLM-Designed Training Environment for RL with Multi-Agent Reasoning}, 
      author={Chao Chen and Chengzu Li and Zhiwei Li and Yinhong Liu and Zhijiang Guo},
      year={2026},
      eprint={2606.17682},
      archivePrefix={arXiv},
      primaryClass={cs.CL},
      url={https://arxiv.org/abs/2606.17682}, 
}

@misc{brumo2025,
  author       = {{Brown University Mathematical Olympiad}},
  title        = {{BRUMO 2025}},
  year         = {2025},
  howpublished = {Problems available at \url{https://huggingface.co/datasets/MathArena/brumo_2025}},
  note         = {Accessed September 2026}
}

@misc{aime2024,
  author       = {{Mathematical Association of America}},
  title        = {{American Invitational Mathematics Examination (AIME) 2024}},
  year         = {2024},
  howpublished = {Problems available at \url{https://huggingface.co/datasets/HuggingFaceH4/aime_2024}},
  note         = {Accessed September 2026}
}

@misc{aime2025,
  author       = {{Mathematical Association of America}},
  title        = {{American Invitational Mathematics Examination (AIME) 2025}},
  year         = {2025},
  howpublished = {Problems available at \url{https://huggingface.co/datasets/math-ai/aime25}},
  note         = {Accessed September 2026}
}

@misc{amc2023,
  author       = {{Mathematical Association of America}},
  title        = {{American Mathematics Competitions (AMC) 12, 2023}},
  year         = {2023},
  howpublished = {Problems available at \url{https://huggingface.co/datasets/AI-MO/aimo-validation-amc}},
  note         = {Accessed September 2026}
}

@misc{hmmt2025feb,
  author       = {{Harvard-MIT Mathematics Tournament}},
  title        = {{HMMT February 2025}},
  year         = {2025},
  howpublished = {\url{https://www.hmmt.org/}; problems available at \url{https://huggingface.co/datasets/MathArena/hmmt_feb_2025}},
  note         = {Accessed September 2026}
}

@misc{fang2026llmzerodiscoveringadaptivetraining,
      title={LLMZero: Discovering Adaptive Training Strategies for RL Post-Training via LLM Agents}, 
      author={Haoyang Fang and Wei Zhu and Boran Han and Alex Zhang and Zhenyu Pan and Shuo Yang and Shuai Zhang and Jiading Gai and Peng Tang and Cuixiong Hu and Xuan Zhu and Huzefa Rangwala and George Karypis and Bernie Wang},
      year={2026},
      eprint={2606.18388},
      archivePrefix={arXiv},
      primaryClass={cs.LG},
      url={https://arxiv.org/abs/2606.18388}, 
}

@misc{karpathy2026autoresearch,
  author       = {Andrej Karpathy},
  title        = {autoresearch: {AI} agents running research on single-{GPU} nanochat training automatically},
  year         = {2026},
  howpublished = {\url{https://github.com/karpathy/autoresearch}},
  note         = {GitHub repository}
}

@misc{he2026reskillreconcilingskillcreation,
      title={ReSkill: Reconciling Skill Creation with Policy Optimization in Agentic RL}, 
      author={Zelin He and Haotian Lin and Boran Han and Wei Zhu and Haoyang Fang and Bernie Wang and Xuan Zhu and Runze Li and Matthew Reimherr},
      year={2026},
      eprint={2606.01619},
      archivePrefix={arXiv},
      primaryClass={cs.AI},
      url={https://arxiv.org/abs/2606.01619}, 
}

@misc{liu2026drkernelreinforcementlearning,
      title={Dr. Kernel: Reinforcement Learning Done Right for Triton Kernel Generations}, 
      author={Wei Liu and Jiawei Xu and Yingru Li and Longtao Zheng and Tianjian Li and Qian Liu and Junxian He},
      year={2026},
      eprint={2602.05885},
      archivePrefix={arXiv},
      primaryClass={cs.LG},
      url={https://arxiv.org/abs/2602.05885}, 
}

@misc{rank2026posttrainbenchllmagentsautomate,
      title={PostTrainBench: Can LLM Agents Automate LLM Post-Training?}, 
      author={Ben Rank and Hardik Bhatnagar and Ameya Prabhu and Shira Eisenberg and Karina Nguyen and Matthias Bethge and Maksym Andriushchenko},
      year={2026},
      eprint={2603.08640},
      archivePrefix={arXiv},
      primaryClass={cs.SE},
      url={https://arxiv.org/abs/2603.08640}, 
}

@misc{lu2024aiscientistfullyautomated,
      title={The AI Scientist: Towards Fully Automated Open-Ended Scientific Discovery}, 
      author={Chris Lu and Cong Lu and Robert Tjarko Lange and Jakob Foerster and Jeff Clune and David Ha},
      year={2024},
      eprint={2408.06292},
      archivePrefix={arXiv},
      primaryClass={cs.AI},
      url={https://arxiv.org/abs/2408.06292}, 
}

@misc{liu2025largelanguagemodelagent,
      title={Large Language Model Agent for Hyper-Parameter Optimization}, 
      author={Siyi Liu and Chen Gao and Yong Li},
      year={2025},
      eprint={2402.01881},
      archivePrefix={arXiv},
      primaryClass={cs.LG},
      url={https://arxiv.org/abs/2402.01881}, 
}

@misc{ma2024eurekahumanlevelrewarddesign,
      title={Eureka: Human-Level Reward Design via Coding Large Language Models}, 
      author={Yecheng Jason Ma and William Liang and Guanzhi Wang and De-An Huang and Osbert Bastani and Dinesh Jayaraman and Yuke Zhu and Linxi Fan and Anima Anandkumar},
      year={2024},
      eprint={2310.12931},
      archivePrefix={arXiv},
      primaryClass={cs.RO},
      url={https://arxiv.org/abs/2310.12931}, 
}

@misc{zhang2025cudaforgeagentframeworkhardware,
      title={CudaForge: An Agent Framework with Hardware Feedback for CUDA Kernel Optimization}, 
      author={Zijian Zhang and Rong Wang and Shiyang Li and Yuebo Luo and Mingyi Hong and Caiwen Ding},
      year={2025},
      eprint={2511.01884},
      archivePrefix={arXiv},
      primaryClass={cs.LG},
      url={https://arxiv.org/abs/2511.01884}, 
}

@misc{yamada2025aiscientistv2workshoplevelautomated,
      title={The AI Scientist-v2: Workshop-Level Automated Scientific Discovery via Agentic Tree Search}, 
      author={Yutaro Yamada and Robert Tjarko Lange and Cong Lu and Shengran Hu and Chris Lu and Jakob Foerster and Jeff Clune and David Ha},
      year={2025},
      eprint={2504.08066},
      archivePrefix={arXiv},
      primaryClass={cs.AI},
      url={https://arxiv.org/abs/2504.08066}, 
}

@article{liu2025understanding,
  title   = {Understanding {R1-Zero}-Like Training: A Critical Perspective},
  author  = {Liu, Zichen and Chen, Changyu and Li, Wenjun and Qi, Penghui and Pang, Tianyu and Du, Chao and Lee, Wee Sun and Lin, Min},
  journal = {arXiv preprint arXiv:2503.20783},
  year    = {2025}
}

@misc{zhang2025stephint,
  title         = {{StepHint}: Multi-level Stepwise Hints Enhance Reinforcement Learning to Reason},
  author        = {Zhang, Kaiyi and Lv, Ang and Li, Jinpeng and Wang, Yongbo and Wang, Feng and Hu, Haoyuan and Yan, Rui},
  year          = {2025},
  eprint        = {2507.02841},
  archivePrefix = {arXiv},
  primaryClass  = {cs.AI},
  url           = {https://arxiv.org/abs/2507.02841}
}

@misc{qu2026pope,
  title         = {{POPE}: Learning to Reason on Hard Problems via Privileged On-Policy Exploration},
  author        = {Qu, Yuxiao and Setlur, Amrith and Smith, Virginia and Salakhutdinov, Ruslan and Kumar, Aviral},
  year          = {2026},
  eprint        = {2601.18779},
  archivePrefix = {arXiv},
  primaryClass  = {cs.LG},
  url           = {https://arxiv.org/abs/2601.18779}
}

@misc{shen2026slim,
  title         = {Dynamic Skill Lifecycle Management for Agentic Reinforcement Learning},
  author        = {Shen, Junhao and Zhang, Teng and Zhao, Xiaoyan and Cheng, Hong},
  year          = {2026},
  eprint        = {2605.10923},
  archivePrefix = {arXiv},
  primaryClass  = {cs.LG},
  url           = {https://arxiv.org/abs/2605.10923}
}
\bibliographystyle{plainnat}

\appendix

\section{Theoretical Analysis}
\label{sec:theory}

We analyze how injecting a scaffold changes the learning signal that GRPO receives, which motivates the design of AutoScaffold. Following~\citet{liao2026selfhintinglanguagemodelsenhance}, we consider binary rewards $R(x, \tau) \in \{0, 1\}$.

\paragraph{Setup.}
For a prompt $x$, GRPO samples a group of $G$ rollouts $\{\tau_i\}_{i=1}^G$ from the current policy $\pi_\theta$ and computes standardized advantages
\begin{equation}
\label{eq:advantage}
A_i = \frac{R_i - \bar{R}}{s + \epsilon}, \quad \bar{R} = \frac{1}{G}\sum_{i=1}^G R_i, \quad s^2 = \frac{1}{G}\sum_{i=1}^G (R_i - \bar{R})^2,
\end{equation}
where $\epsilon \geq 0$ is a numerical stabilizer. The policy gradient is weighted by these advantages: when $A_i = 0$ for all $i$, the group contributes no gradient regardless of the rollout content.

\paragraph{Degenerate groups and the learning cliff.}
Under binary rewards, each rollout independently succeeds with probability $p = p_\theta(x)$. A group is \emph{degenerate} if all rollouts share the same outcome ($\bar{R} \in \{0, 1\}$), in which case $s = 0$ and $A_i = 0$ for all $i$. This underlies the \emph{learning cliff}~\citep{zhang2026scafgrposcaffoldedgrouprelative,li2026questaexpandingreasoningcapacity}: when $p$ is very small, nearly all groups are all-failure and the task contributes almost no learning signal. The probability of obtaining a non-degenerate group therefore determines whether a prompt contributes to learning at all.

\begin{proposition}[Non-degenerate group probability~\citep{liao2026selfhintinglanguagemodelsenhance}]
\label{prop:gate}
For a prompt with success probability $p \in [0, 1]$ and group size $G \geq 2$,
\begin{equation}
\label{eq:gate-app}
u(p) := \Pr[s > 0] = 1 - (1 - p)^G - p^G.
\end{equation}
$u$ is strictly concave on $[0,1]$ and symmetric about $p=\tfrac{1}{2}$, its unique maximizer, with $u(\tfrac{1}{2})=1-2^{1-G}$. Hence $u$ is strictly increasing on $[0,\tfrac{1}{2}]$ and strictly decreasing on $[\tfrac{1}{2},1]$, and $u(p) = Gp + O(p^2)$ as $p \to 0$.
\end{proposition}

\begin{proof}
A group is degenerate exactly when all $G$ rollouts fail or all succeed, which occurs with probability $(1-p)^G + p^G$. Moreover, $u''(p) = -G(G-1)\bigl[p^{G-2} + (1-p)^{G-2}\bigr] < 0$ for all $p \in [0,1]$, so $u$ is strictly concave. Together with the symmetry $u(p) = u(1-p)$, this implies that $u$ has a unique maximum at $p = \tfrac{1}{2}$.
\end{proof}

When $p \ll 1$, which is the common regime for hard tasks, $u(p) \approx Gp$: the group is non-degenerate only when at least one rollout succeeds, an event that becomes rare as $p \to 0$. Conversely, when $p \approx 1$, $u(p) \approx G(1-p)$: almost all groups are all-success, again yielding no gradient.

\begin{proposition}[Expected within-group reward variance~\citep{liao2026selfhintinglanguagemodelsenhance}]
\label{prop:energy}
Define the within-group reward variance $V = \frac{1}{G}\sum_{i=1}^G (R_i - \bar{R})^2$. Under binary rewards with success probability $p$,
\begin{equation}
\label{eq:energy}
\mathbb{E}[V \mid x] = \frac{G-1}{G}\, p(1-p),
\end{equation}
which is strictly concave in $p$ and uniquely maximized at $p = \tfrac{1}{2}$.
\end{proposition}

\begin{proof}
For binary rewards, $V = \bar{R}(1 - \bar{R})$. Since $G\bar{R} \sim \mathrm{Binomial}(G, p)$, we have $\mathbb{E}[\bar{R}] = p$ and $\mathbb{E}[\bar{R}^2] = p^2 + p(1-p)/G$. Hence $\mathbb{E}[V \mid x] = p - p^2 - p(1-p)/G = \frac{G-1}{G}\,p(1-p)$.
\end{proof}

Propositions~\ref{prop:gate} and~\ref{prop:energy} show that $p=\tfrac{1}{2}$ maximizes both the probability of a non-degenerate group and its expected reward variance. The two quantities play different roles under GRPO. Because advantages are standardized, any non-degenerate group yields advantages of roughly unit scale ($\frac{1}{G}\sum_i A_i^2 = s^2/(s+\epsilon)^2 \approx 1$), so the primary quantity is $u(p)$, i.e., whether a prompt contributes at all. The reward variance measures signal strength when advantages are not normalized by $s$, and we use it below as a tractable proxy for class-level analysis. Both quantities are symmetric and concave, vanish at $p\in\{0,1\}$, and peak at $p=\tfrac{1}{2}$, supporting the same qualitative conclusions.

\paragraph{Implications for scaffold design.}
Let $p_\theta(x)$ denote the student's success probability without scaffold and $p_\theta(x, S)$ the probability under scaffold $S$. A scaffold that raises $p_\theta(x, S)$ from near zero toward $\tfrac{1}{2}$ increases both the non-degenerate group probability and the expected reward variance, enabling gradient updates on tasks that would otherwise contribute almost nothing. A scaffold that is too strong instead overshoots: if $p_\theta(x, S) \approx 1$, groups become degenerate again (all-success), and the learning signal collapses despite a high success rate. An effective scaffold should therefore make success reachable while leaving meaningful work for the student, for example by providing strategies or partial hints rather than complete solutions.

\begin{remark}[Scaffold should adapt to the student]
\label{rem:adaptive}
The per-step signal is largest when $p_\theta(x, S) \approx \tfrac{1}{2}$. Since $p_\theta(x, S)$ depends on $\theta$, a scaffold that brings a task close to this point at one stage of training generally does not keep doing so as the student changes: as the student improves, the same scaffold may push $p_\theta(x, S)$ toward $1$, and as its failures shift, the scaffold may no longer address them. Hence no fixed scaffold keeps every task in the high-signal regime throughout training~\citep[cf.][Proposition~3.3]{liao2026selfhintinglanguagemodelsenhance}. This statement concerns the per-step signal and does not by itself characterize the long-term learning trajectory.
\end{remark}

\paragraph{Class-level scaffold and aggregate learning signal.}
The preceding analysis concerns a single prompt with success probability $p$. Per-task success rates can only be estimated from $G$ rollouts and are therefore noisy, so we analyze scaffolds at the class level, in terms of the class success rates without scaffold ($s_C$) and with scaffold ($\sigma_C$).

Consider a task $x$ drawn uniformly from class $C$. Its success probability $p_x$ varies across tasks, reflecting the heterogeneous difficulty of the class. We write $p_x = s_C + \delta_x$, where $\delta_x$ is a zero-mean perturbation with $\mathrm{Var}(\delta_x) = v_C$. The class-average within-group reward variance is
\begin{equation}
\label{eq:class_signal}
\bar{V}_C := \frac{G-1}{G}\,\mathbb{E}_x\bigl[p_x(1-p_x)\bigr] = \frac{G-1}{G}\bigl[s_C(1-s_C) - v_C\bigr],
\end{equation}
where the cross term vanishes because $\mathbb{E}[\delta_x] = 0$. The term $s_C(1-s_C)$ corresponds to a homogeneous class with uniform difficulty $s_C$, and the subtracted $v_C$ is the reduction due to task heterogeneity (consistent with Jensen's inequality, since $p(1-p)$ is concave).

\begin{assumption}[Uniform shift]
\label{assump:shift}
A class-level scaffold shifts every task equally: $p_x^S=p_x+(\sigma_C-s_C)\in[0,1]$ for $x\in C$, preserving variance $v_C$.
\end{assumption}

Assumption~\ref{assump:shift} is an idealization. In practice, a shared strategy may help some tasks more than others and may leave the hardest tasks at zero success; we treat such tasks separately below.

\begin{proposition}[Class-level scaffold benefit]
\label{prop:class}
For fixed $v_C$ and $s_C\in[0,\tfrac{1}{2})$, $\bar{V}_C$ increases with $s_C$:
\begin{equation}
\frac{\partial \bar{V}_C}{\partial s_C} = \frac{G-1}{G}(1 - 2s_C) > 0.
\end{equation}
Under Assumption~\ref{assump:shift}, if a class-level scaffold raises the class success rate from $s_C$ to $\sigma_C$, the change in class-average reward variance is
\begin{equation}
\label{eq:class_improvement}
\bar{V}_C^S - \bar{V}_C = \frac{G-1}{G}\,(\sigma_C - s_C)(1 - \sigma_C - s_C),
\end{equation}
which is positive if and only if $s_C < \sigma_C < 1 - s_C$.
\end{proposition}

\begin{proof}
The derivative follows from Eq.~(\ref{eq:class_signal}). Under Assumption~\ref{assump:shift}, scaffolding gives mean $\sigma_C$ and unchanged variance $v_C$, so $\bar{V}_C^S = \frac{G-1}{G}\bigl[\sigma_C(1-\sigma_C) - v_C\bigr]$. Subtracting gives
\begin{equation*}
\bar{V}_C^S - \bar{V}_C = \frac{G-1}{G}\bigl[\sigma_C(1-\sigma_C) - s_C(1-s_C)\bigr] = \frac{G-1}{G}\,(\sigma_C - s_C)(1 - \sigma_C - s_C).
\end{equation*}
The first factor is positive if and only if $\sigma_C > s_C$, and the second if and only if $\sigma_C < 1 - s_C$.
\end{proof}

The improvement depends only on the two class-level rates $s_C$ and $\sigma_C$. Writing $\Delta_C = \sigma_C - s_C$ for the scaffold effect, the improvement $\frac{G-1}{G}\Delta_C(1 - 2s_C - \Delta_C)$ is larger for classes with lower $s_C$ at a fixed effect, and it is negative for every $\Delta_C > 0$ once $s_C \ge \tfrac{1}{2}$: a scaffold that raises the success rate of a class already above one half reduces the class-level signal. It is proportional to both the scaffold effect $(\sigma_C - s_C)$ and the factor $(1 - \sigma_C - s_C)$, and for a given $s_C$ it is largest at $\sigma_C = \tfrac{1}{2}$. If the scaffold is so strong that $\sigma_C \geq 1 - s_C$, the class-level signal no longer increases and can even decrease, although the success rate is higher. This formalizes why class-level scaffolds should raise success rates toward, but not far beyond, $\tfrac{1}{2}$.

\paragraph{Instance-level scaffold and zero-success tasks.}
Class-level scaffolds mainly benefit tasks that the student can already partially solve ($p_x > 0$), lifting their success rates toward the high-signal regime. For tasks where the student's success probability is zero, a shared strategy is often insufficient, and targeted support such as a partial solution or an intermediate result is needed.

Let $\mathcal{Z}_C = \{x \in C : p_\theta(x) = 0\}$ denote the set of zero-success tasks in class $C$, with $\alpha_C = |\mathcal{Z}_C| / |C|$. These tasks produce only all-failure groups, contributing no reward-relative gradient or within-group reward variance:
\begin{equation}
\label{eq:zero_partition}
\bar{V}_C = \frac{G-1}{G}\,(1 - \alpha_C)\,\mathbb{E}_{x \notin \mathcal{Z}_C}\bigl[p_x(1-p_x)\bigr].
\end{equation}

\begin{proposition}[Instance-level scaffold for zero-success tasks]
\label{prop:instance}
Suppose an instance-level scaffold $S_{\mathrm{inst}}[x]$ is applied to each zero-success task $x \in \mathcal{Z}_C$, raising its success probability from $0$ to $p_x^S \in (0, 1)$, while the remaining tasks are unchanged. The class-average variance increases by
\begin{equation}
\label{eq:instance_improvement}
\bar{V}_C^S - \bar{V}_C = \frac{G-1}{G}\,\alpha_C\,\mathbb{E}_{x \in \mathcal{Z}_C}\bigl[p_x^S(1-p_x^S)\bigr] > 0.
\end{equation}
\end{proposition}

\begin{proof}
With instance-level scaffolds, the class-average reward variance becomes
$\bar{V}_C^S = \frac{G-1}{G}\bigl[(1-\alpha_C)\,\mathbb{E}_{x \notin \mathcal{Z}_C}[p_x(1-p_x)] + \alpha_C\,\mathbb{E}_{x \in \mathcal{Z}_C}[p_x^S(1-p_x^S)]\bigr]$.
Subtracting Eq.~(\ref{eq:zero_partition}) yields Eq.~(\ref{eq:instance_improvement}), which is positive because $p_x^S(1-p_x^S) > 0$ for $p_x^S \in (0, 1)$.
\end{proof}

The improvement is proportional to $\alpha_C$: the more zero-success tasks there are, the greater the benefit, which is particularly relevant early in training. The requirement $p_x^S < 1$ matters: a hint that reveals the complete solution makes every rollout succeed, so the group is again degenerate and contributes nothing. Propositions~\ref{prop:class} and~\ref{prop:instance} thus formalize the complementary roles of the two scaffold levels: class-level scaffolds shift the difficulty of tasks the student can already attempt toward the high-signal regime, while instance-level scaffolds expand the set of tasks that contribute to training at all. This motivates using instance-level scaffolds mainly for tasks that remain at zero success under class-level scaffold, with hints that are partial rather than complete.

\paragraph{Bounded rewards.}
The analysis extends to rewards in $[0,1]$, such as the rubric scores used for KernelBench. A group yields zero advantage whenever all its rewards are equal, so for a reward taking values $v$ with probabilities $\pi_v$, the probability of a non-degenerate group is $1-\sum_v \pi_v^G$, which is small whenever one value dominates, as the zero reward does on hard tasks. The expected within-group reward variance is $\mathbb{E}[V \mid x] = \frac{G-1}{G}\,\mathrm{Var}(R \mid x)$, and for a reward in $[0,1]$ with mean $\mu$, $\mathrm{Var}(R \mid x) \le \mu(1-\mu)$, with equality for binary rewards. Both quantities therefore vanish when nearly every rollout receives the same reward, and a scaffold helps in the same way: by making rewards within a group differ.

\paragraph{Scope of the analysis.}
The analysis above characterizes the learning signal available under the scaffolded prompt, i.e., the signal for updating $\pi_\theta(\cdot \mid x, S)$. It does not model how improvements under scaffolds transfer to the unscaffolded policy $\pi_\theta(\cdot \mid x)$, which is what we evaluate. In particular, with injection probability $q$, the expected class-level signal $(1-q)\,\bar{V}_C(s_C) + q\,\bar{V}_C(\sigma_C)$ is linear in $q$ and would favor $q = 1$ whenever the scaffold helps. In practice, an always-on scaffold creates a mismatch between training and evaluation and encourages reliance on privileged information (cf.\ the $q_{\max}{=}1$ ablation in Section~\ref{sec:ablation}), which motivates capping $q$ at $q_{\max} < 1$. Characterizing transfer remains future work.

\section{Case Study: How the Teacher Gathers Evidence and Decides}
\label{app:case_studies}

\subsection{ALFWorld}
\label{app:case:alfworld}

This section gives the full record of one teacher update on ALFWorld, at step 30 of a separate training run with the same setting, taken from the teacher's decision log. The tool returns are rebuilt from the logged training rollouts with the same tool code, and their lengths match the logged lengths exactly.

\paragraph{Task classes.}
ALFWorld tasks are played by issuing text commands such as \texttt{go to microwave 1}, \texttt{take mug 1 from countertop 1}, or \texttt{heat mug 1 with microwave 1}. The classes involved in this update are Heat (heat an object, which is possible only in the microwave, then put it somewhere), Look (examine an object under a lamp, which requires turning the lamp on with \texttt{use}), Pick2 (put two objects of the same kind in one place), Clean (clean an object at the sink, then put it somewhere), and Cool (cool an object in the fridge, then put it somewhere). Each training game is attempted 8 times, and a game in which all 8 attempts fail gives no learning signal.

\paragraph{Scaffold before the update.}
Heat, Look, and Clean each have a class-level scaffold, with injection probabilities 0.45, 0.4, and 0.4; the other classes have none.

\paragraph{Initial evidence.}
The update starts from two evaluations on the held-out tasks $V_C$. Without a scaffold, the student solves 27.3\% of the tasks: 6.2\% of Heat, 19.6\% of Look, 21.6\% of Clean, 26.5\% of Cool, 24.2\% of Pick2, and 47.5\% of Pick. On 20 games per class, evaluated with and without the current scaffold, success with the scaffold is 55\% versus 0\% on Heat, 40\% versus 10\% on Clean, and 45\% versus 35\% on Look.

\paragraph{Gathering evidence.}
The teacher made six tool calls, in this order:
\begin{enumerate}
\item $Q_{\text{stat}}$ (training statistics of the current block). Heat attempts succeed 3\% of the time without a scaffold and 27\% with it, and seven of eight Heat games without a scaffold fail on every attempt. In five of eight Look games that received the scaffold, every attempt still fails. In 30 of 39 Pick2 games, every attempt fails.
\item Memory $M$ (records of past updates). In the previous update, attempts with a scaffold already did better than those without on Look (23\% versus 14\%) and Heat (18\% versus 4\%).
\item[3--6.] $Q_{\text{traj}}$, four failed attempts from each of Pick2, Cool, Heat, and Look (the first 24 steps of each):
\begin{itemize}
\item Pick2: the student puts the first object down and picks it up again, and once tries to move both objects with one command, \texttt{put both spraybottles into garbagecan 1}, which the environment rejects.
\item Cool: the student keeps putting the pot back on the stove and never goes to the fridge.
\item Heat: 38 actions go to or use the coffee machine or the stove; none involve the microwave.
\item Look: the student repeats \texttt{examine cellphone 1 with desklamp}, which the environment rejects, and never issues \texttt{use}, so the lamp is never turned on.
\end{itemize}
\end{enumerate}

\paragraph{Diagnosis.}
The teacher's diagnosis, which it wrote before proposing edits, makes five points:
\begin{itemize}
\item \textbf{Heat} is the weakest class without a scaffold (6.2\% on $V_C$; seven of eight training games without a scaffold fail on every attempt), while the current scaffold gives a large gain (55\% versus 0\%). The teacher therefore keeps showing Heat a scaffold often, but replaces its text with one written specifically for heating.
\item \textbf{Clean} still benefits from its scaffold (34\% versus 14\% in training, 40\% versus 10\% on $V_C$), so the teacher keeps it.
\item \textbf{Pick2} has the most games in which every attempt fails (30 of 39), and its failed attempts place only one object or invent a command that moves both objects at once, so the teacher proposes a step-by-step procedure for placing two objects.
\item \textbf{Look} attempts repeat the invalid command \texttt{examine <object> with desklamp} and never use the lamp, so the teacher proposes a scaffold about turning the lamp on.
\item \textbf{Cool} is improving without a scaffold, so the teacher leaves it without one for now.
\end{itemize}

\paragraph{Proposed edits.}
The teacher proposed new class-level text for each of the three failures it found, rewriting the current Heat and Look scaffolds and adding one for Pick2:
\begin{itemize}
\item \textbf{Heat} ($q=0.5$): \textit{``For heat tasks, do not use the coffeemachine or stoveburner. Find and take the exact target object, go to microwave 1, then run `heat <object> with microwave 1'. After it is heated, carry that same object to the requested receptacle and put it in/on there once.''}
\item \textbf{Look} ($q=0.4$): \textit{``For look-in-light tasks, first find and take the exact target object. Then go to a lamp such as desklamp 1 or floorlamp 1 and use the lamp with `use <lamp>'. Do not keep trying `examine <object> with desklamp'; the important final step is to be carrying the object at the lamp and use the lamp.''}
\item \textbf{Pick2} ($q=0.45$): \textit{``For two-object tasks, place two separate objects with the requested name in the same receptacle. Do not use `both' or a plural command. Take one target object, go to the receptacle, put it in/on there; then find a different second target object, take it, and put it in/on that same receptacle.''}
\end{itemize}
It kept the Clean scaffold and wrote nothing for Cool.

\paragraph{Validation and decision.}
Each proposed scaffold was validated on 30 games per class from $V_C$ (Table~\ref{tab:case-ab}). The Heat scaffold improves on the current text (40.0\% versus 30.0\%), and the Look scaffold matches the current one, so both are adopted. The Pick2 scaffold measures below the no-scaffold condition (10.0\% versus 13.3\%), so the teacher drops it and leaves Pick2 without a scaffold. The final update replaces the Heat and Look scaffolds with the new text, keeps the Clean scaffold, and sets $q$ to 0.5 for Heat, 0.4 for Look, 0.45 for Clean, and 0 for the other classes.

\begin{table}[h]
\caption{Validation of the scaffolds proposed in this update: success rate (\%) on 30 held-out games per class. Pick2 has no current scaffold.}
\label{tab:case-ab}
\centering
\small
\begin{tabular}{lcccl}
\toprule
Class & No scaffold & Current scaffold & Proposed scaffold & Outcome \\
\midrule
Heat & 6.7 & 30.0 & 40.0 & adopted \\
Look & 26.7 & 30.0 & 30.0 & adopted \\
Pick2 & 13.3 & -- & 10.0 & dropped \\
\bottomrule
\end{tabular}
\end{table}

\paragraph{Outcome.}
In the next block, attempts with the new scaffolds succeed far more often than attempts without them on Look (67\% versus 4\%) and Heat (51\% versus 16\%). At the next update, success without a scaffold on $V_C$ has risen from 27.3\% to 36.7\% overall, from 6.2\% to 16.7\% on Heat, and from 19.6\% to 34.7\% on Look. Clean, whose scaffold was not changed, also rises, from 21.6\% to 47.4\%, so part of these gains comes from RL training itself. The Look and Heat scaffolds are removed three and four updates later, once they no longer help the student.

\subsection{Mathematical Reasoning}
\label{app:case:math}

This section gives the full record of one teacher update in math, at step 100 of the training run shown in Figure~\ref{fig:math_ratio}, where the mean prefix ratio stops rising and begins to fall. The record is taken from the teacher's decision log and the run's per-update statistics. The log stores which tools the teacher called and in what order, its written diagnosis, and its proposed edits; it does not store the tool arguments or the returned contents, so we report the statistics the teacher had access to at this update rather than the exact tool returns.

\paragraph{Setting.}
The student is Nemotron-1.5B (OpenMath-Nemotron-1.5B), trained on the 12{,}506 problems released with QuestA. The instance-level scaffold follows QuestA~\citep{li2026questaexpandingreasoningcapacity}: the first $r_x$ percent of a reference solution is prepended to problem $x$, with $r_x \in [0, 50]$. Each problem starts at the ratio its QuestA difficulty filter selected it for: 1{,}853 problems at $r_x = 50$ and 10{,}653 at $r_x = 25$, so no problem starts without a hint. Each update covers one block of 10 training steps, in which 2{,}560 problems are each attempted 8 times. A problem whose 8 attempts all fail, or all succeed, gives no learning signal. The teacher can change $r_x$ for groups of problems selected by their outcome in the current block (all-fail, all-pass, or mixed), by their current $r_x$, and by their success rate. Math has a single task class (Table~\ref{tab:classes}), so its class-level scaffold consists of short notes that are added to a training prompt with injection probability $q$, with at most 8 notes in force at a time. A problem that is solved with $r_x = 0$ is marked as graduated.

\paragraph{Scaffold before the update.}
Across the pool, the mean prefix ratio is 33.7. About 3{,}500 problems are at $r_x = 25$ and about 4{,}200 at $r_x = 50$, and the rest are at intermediate ratios; 130 problems have been reduced to $r_x = 0$, and 46 have graduated. Eight class-level notes are in force with $q = 0.2$. They tell the student to check that the boxed answer is the quantity the problem asks for; to box the option letter in multiple-choice problems; not to stop after finding an intermediate result; to use a contingency table for two-attribute counts; to respect the order of indexed variables in extremal problems; to use direction vectors when counting lines and planes; to read ``sufficient'' and ``necessary'' as implications; and to follow inequality chains by index residue.

\paragraph{Initial evidence.}
In the current block, 575 of the 2{,}560 problems are all-fail, 1{,}829 are mixed, and 156 are all-pass, so 28.6\% of the problems (the all-fail and all-pass ones) give no learning signal. Among problems that received the notes, 21.6\% are all-fail, compared with 22.7\% among problems that did not. The teacher also has the student's hint-free success on the held-out tasks $V_C$ (200 problems), which it uses to check that the student is not becoming dependent on hints.

\paragraph{Gathering evidence.}
The teacher made five tool calls, in this order:
\begin{enumerate}
\item $Q_{\text{stat}}$ (training statistics of the current block): the number of all-fail, mixed, and all-pass problems at each prefix ratio, and the outcomes of earlier ratio changes on problems seen again since (revisits).
\item[2--4.] $Q_{\text{stat}}$ three more times, to list individual problems (the selection arguments are not recorded).
\item[5.] $Q_{\text{traj}}$, to read failed attempts on all-fail problems.
\end{enumerate}

\paragraph{Diagnosis.}
The teacher's diagnosis, written before it proposed edits (in its notes, $r$ is the prefix ratio $r_x$, ``probes'' are the hint-free evaluations on $V_C$, and ``text probability'' is $q$):
\begin{quote}\small
\textit{``Cycle 10 still has a large zero-gradient bottleneck: 575/2560 groups are all-fail. Most are already high-dose, but the 119 all-fail problems at $r \le 25$ still have useful headroom; revisit evidence says raising prior all\_fail@25 to r45/r50 escaped about 55--60\%, so those should be boosted. For all-fail problems at intermediate/high r with remaining headroom, raising toward 50 is still justified; prior all\_fail@45 $\to$ r50 escaped about 39\%. Problems already mixed have gradient, so follow the standing slow-wean rule for succ$>$50\%; all-pass groups can be annealed faster. Hint-free probes are not clearly deteriorating overall, so keep text probability unchanged, but add a couple of concise general skills aimed at observed max-dose failures where ratio cannot help: two-way table/conditional-count discipline and geometric halving processes.''}
\end{quote}
The diagnosis makes four points:
\begin{itemize}
\item \textbf{All-fail problems with room to raise $r_x$.} Of the 575 all-fail problems, 119 have $r_x \le 25$. When the teacher earlier raised all-fail problems from $r_x = 25$ to 45 or 50, about 55--60\% of them were no longer all-fail when seen again, so the teacher raises these problems again.
\item \textbf{All-fail problems at a high ratio.} Most all-fail problems are already at or near $r_x = 50$. For those below 50, raising $r_x$ still helped in the past (about 39\% of problems raised from 45 to 50 escaped all-fail), so the teacher raises them toward 50. For those already at 50, a larger prefix is not possible.
\item \textbf{Mixed and all-pass problems.} Mixed problems already give a learning signal, so the teacher only reduces $r_x$ slowly, for problems solved in more than half of their attempts. All-pass problems are reduced faster.
\item \textbf{Notes.} Hint-free success on $V_C$ is not clearly falling, so the teacher keeps $q$ unchanged. For problems that fail at the maximum ratio, where a longer prefix cannot help, it proposes two new notes based on failures it read: one on two-way tables and conditional counts, and one on processes that halve a quantity at each stage.
\end{itemize}

\paragraph{Decision.}
The teacher made adjustments in opposite directions for different groups of problems (Table~\ref{tab:math-case-ops}). It proposed two notes:
\begin{itemize}
\item \textit{``For problems with two attributes or categories, make a 2x2 table. Distinguish marginal counts from conditional counts such as `among the clayey loam'; subtract only within the same row/column, then add the cells the question asks for.''}
\item \textit{``When a process halves the remaining amount at each stage, successive equal-fraction stages usually take half as long as the previous one. Use a geometric sum for total time/amount, not a constant time per stage.''}
\end{itemize}
It left $q$ at 0.2. All-fail problems at $r_x = 50$ were left unchanged.

\begin{table}[h]
\caption{Prefix-ratio changes in this update. Ratios are capped at 50.}
\label{tab:math-case-ops}
\centering
\small
\begin{tabular}{llr}
\toprule
Outcome in the current block & Selection & Change in $r_x$ \\
\midrule
All-fail & $0 \le r_x \le 25$ & $+20$ \\
All-fail & $26 \le r_x \le 49$ & $+10$ \\
Mixed & $r_x \ge 1$, success rate $> 50\%$ & $-5$ \\
All-pass & $r_x \ge 1$ & $-10$ \\
\bottomrule
\end{tabular}
\end{table}

\paragraph{Outcome.}
The four ratio changes were applied. The two notes were not: eight notes were already in force, the maximum, and the framework rejects note additions that would exceed it, so the notes in force stayed the same. Over the next three updates, the number of problems the teacher had reduced to $r_x = 0$ grew from 130 to 177, 244, and 303, the number of graduated problems rose from 46 to 90, all-pass problems in each block rose from 156 to 241, and the mean prefix ratio fell from 33.7 to 33.3. At the next update the teacher reported, from revisits, that raising all-fail problems from $r_x = 25$ to 50 had let 86 of 117 escape all-fail, and that cuts of about 10 points on all-pass problems were mostly safe, while larger cuts at high ratios sometimes turned a problem back to all-fail. Hint-free pass@1 on $V_C$ rose from 28.8\% at step 80 to 34.1\% at step 120 and 38.3\% at step 160.

As in ALFWorld, the update follows the student's current ability: where more of the same support can still help, the teacher gives more of it; where the student already succeeds, it gives less; and where the support cannot be increased further, it proposes a different kind of support.

\section{Evaluation Domains}
\label{app:domains}

We evaluate \textsc{AutoScaffold} on four domains that differ in task structure, interaction mode, and reward.
\paragraph{ALFWorld.}
ALFWorld~\citep{shridhar2021alfworldaligningtextembodied} is a text-based household environment in which the agent completes long-horizon tasks through multi-step planning and object interaction. It defines six task types, each used as a task class for class-level scaffolds: Pick \& Place (Pick), Examine in Light (Look), Clean \& Place (Clean), Heat \& Place (Heat), Cool \& Place (Cool), and Pick Two \& Place (Pick2). We evaluate on the standard 140-game test split; an episode succeeds if the task is completed within the environment's step limit, and we report the success rate.

\paragraph{Search-based QA.}
The model answers open-domain questions by interacting with a search tool, deciding when and what to query and how to combine the retrieved passages into a final answer. We train on NQ and HotpotQA and evaluate on three single-hop datasets, NQ~\citep{kwiatkowski-etal-2019-natural}, TriviaQA~\citep{joshi2017triviaqalargescaledistantly}, and PopQA~\citep{mallen2023trustlanguagemodelsinvestigating}, and four multi-hop datasets, HotpotQA~\citep{yang2018hotpotqadatasetdiverseexplainable}, 2Wiki~\citep{ho2020constructingmultihopqadataset}, MuSiQue~\citep{trivedi2022musiquemultihopquestionssinglehop}, and Bamboogle~\citep{press2023measuringnarrowingcompositionalitygap}. NQ and HotpotQA are in-domain; the others are out-of-domain. Exact match (EM) is used as the reward and the evaluation metric.

\paragraph{KernelBench.}
KernelBench~\citep{ouyang2025kernelbenchllmswriteefficient} asks the model to write GPU kernels that reproduce a reference PyTorch implementation and run faster than it. Level~1 contains 100 relatively simple tasks involving basic operators (e.g., matrix multiplication); Level~2 contains 100 medium-difficulty tasks composed of multi-step operator combinations; Level~3 contains 50 challenging tasks involving full network architectures (e.g., AlexNet). We train on a dataset of 600 kernel generation tasks. Correctness is verified against the reference output, and we report correctness and Fast$_1$, the fraction of tasks with a correct kernel that is faster than the PyTorch reference. Speed is measured on H200.

\paragraph{Math reasoning.}
The student performs multi-step reasoning and outputs a final answer, which is checked against the ground truth. We use eight competition-level benchmarks: the recent contests AIME~24~\citep{aime2024}, AIME~25~\citep{aime2025}, HMMT~Feb~25~\citep{hmmt2025feb}, and BRUMO~25~\citep{brumo2025}, together with AMC~23~\citep{amc2023}, MATH-500~\citep{lightman2023letsverifystepstep}, Minerva Math~\citep{lewkowycz2022solvingquantitativereasoningproblems}, and OlympiadBench~\citep{he2024olympiadbenchchallengingbenchmarkpromoting}. Following the corresponding baselines, the Qwen3-4B-Instruct setting (following SAGE) reports AIME~24/25, AMC~23, MATH-500, Minerva Math, and OlympiadBench (Table~\ref{tab:math-qwen3-4b}), and the Nemotron-1.5B setting (following QuestA) reports AIME~24/25, HMMT~Feb~25, OlympiadBench, and BRUMO~25 (Table~\ref{tab:math-main}). We follow the evaluation settings of each baseline's paper.

\section{Teacher Cost}
\label{app:cost}

\paragraph{Cost per scaffold update.}
Each scaffold update consists of (i) investigation, in which the teacher queries $Q_{\text{stat}}$, $Q_{\text{traj}}$, and $Q_{\text{group}}$ and reads its memory; (ii) proposal; and (iii) held-out validation of class-level edits, which runs student rollouts under three conditions. Teacher calls run on an external API; validation rollouts run on the training GPUs. Table~\ref{tab:teacher-cost} reports the cost per domain. Validation dominates the time per update, while the API cost of the teacher is a few dollars per run.

\begin{table}[h]
\caption{Teacher cost per domain. Per-update values are means over all scaffold updates in a run.}
\label{tab:teacher-cost}
\centering
\small
\begin{tabular}{lcccc}
\toprule
 & ALFWorld & Search QA & Math & KernelBench \\
\midrule
Teacher wall-clock per update (min) & 4 & 3 & 5 & 4 \\
Validation wall-clock per update (min) & 11 & 13 & 21 & 16 \\
Total teacher API cost per run (USD) & 2.35 & 3.13 & 5.57 & 2.76 \\
\bottomrule
\end{tabular}
\end{table}

\end{document}